\documentclass[10pt,twocolumn,letterpaper]{article}

\usepackage{cvpr}
\usepackage{times}
\usepackage{epsfig}
\usepackage{graphicx}
\usepackage{amsmath}
\usepackage{amssymb}
\usepackage[ruled]{algorithm2e}
\usepackage{subcaption}
\usepackage{array}
\usepackage{color}

\definecolor{sea}{RGB}{0,94,154}
\definecolor{tree}{RGB}{125,166,0}
\definecolor{sand}{RGB}{233,142,0}
\definecolor{mountain}{RGB}{0,52,83}
\definecolor{plant}{RGB}{170,192,0}
\definecolor{road}{RGB}{192,210,154}
\definecolor{bridge}{RGB}{91,177,1}
\definecolor{sign}{RGB}{76,244,50}
\definecolor{building}{RGB}{218,86,18}
\definecolor{window}{RGB}{167,147,142}
\definecolor{door}{RGB}{144,23,0}
\definecolor{color12}{RGB}{254,254,0}
\definecolor{color13}{RGB}{127,254,254}
\definecolor{color14}{RGB}{127,0,254}

\makeatletter
\newcommand{\thickhline}{%
    \noalign {\ifnum 0=`}\fi \hrule height 1pt
    \futurelet \reserved@a \@xhline
}
\newcolumntype{"}{@{\hskip\tabcolsep\vrule width 1pt\hskip\tabcolsep}}
\makeatother

\usepackage[pagebackref=true,breaklinks=true,letterpaper=true,colorlinks,bookmarks=false]{hyperref}

 \cvprfinalcopy 

\def\cvprPaperID{2238} 

\ifcvprfinal\fi
\begin{document}

\title{Sample and Filter: Nonparametric Scene Parsing via Efficient Filtering}

\author{Mohammad Najafi${}^{1,2}$ \,\,\,\,\,  Sarah Taghavi Namin${}^{1,2}$ \,\,\,\,\, Mathieu Salzmann${}^{2,3}$ \,\,\,\,\, Lars Petersson${}^{1,2}$\\
${}^{1}$Australian National University (ANU)\,\,\,\,\,\,\,\,\,\,\,\,\,\,\,\,\,${}^{2}$NICTA\thanks{\tiny NICTA is funded by the Australian Government through the Department of Communications and the Australian Research Council through the ICT Centre of Excellence Program.}\,\,\,\,\,\,\,\,\,\,\,\,\,\,\,\,\,${}^{3}$CVLab, EPFL, Switzerland\\
{\tt\small \{sarah.namin, mohammad.najafi,  lars.petersson\}@nicta.com.au\,\,\,\,\,\,\,\,\,\, mathieu.salzmann@epfl.ch}
\vspace {-10pt}
}
\newcommand{\cmt}[1]{\textcolor{red}{\textbf {#1}}}
\newcommand{\comment}[1]{}

\def\model{{\cal M}}
\def\state{\mathbf{\phi}}
\def\twindow{\tau}
\def\thetaX{\ba}
\def\thetaY{\bb}
\def\vecthetaX{\bA}
\def\vecthetaY{\bB}
\def\basisX{\phi}
\def\basisY{\psi}
\def\hpX{\mathbf{\bar{\alpha}}}
\def\hpY{\mathbf{\bar{\beta}}}
\def\bvel{\bv}
\def\Xdim{{d}}
\def\Ydim{{D}}
\newcommand{\Xin}{\bX_{\mathit{in}}}
\newcommand{\Xout}{\bX_{\mathit{out}}}
\newcommand{\Xoutm}{\bX_{\mathit{out,m}}}
\newcommand{\Xoutmbar}{\bar{\bX}_{\mathit{out,m}}}
\def\timestep{{\Delta t}}
\def\tx{\mathbf{\tilde{\bx}}}
\def\ty{\mathbf{\tilde{\by}}}
\def\Xnew{\mathbf{{\bX}^\prime}}
\def\x1new{\mathbf{{\bx}_1^\prime}}
\def\Xoutnew{\mathbf{{\bX}^\prime_{out}}}
\def\y1new{\mathbf{{\by}_1^\prime}}
\def\Youtnew{\mathbf{{\bY}^\prime_{out}}}
\def\Ynew{\mathbf{{\bY}^\prime}}

\def\balpha{\boldsymbol\alpha}
\def\bgamma{\boldsymbol\gamma}

\newcommand{\vt}[1]{{\bf v}_{#1}}
\newcommand{\vtt}[2]{{{\bf v}_{#1}^{#2}}^T}
\newcommand{\Mm}[1]{{\bf M'_m}^{#1}}
\newcommand{\argmin}{\operatornamewithlimits{argmin}}

\newcommand{\tr}[1]{\ensuremath{\mathrm{tr}\left(#1\right)}}
\newcommand{\maximize}{\operatornamewithlimits{maximize}}
\newcommand{\minimize}{\operatornamewithlimits{minimize}}

%
%

\newcommand{\ba}{\mathbf{a}}
\newcommand{\bA}{\mathbf{A}}
\newcommand{\bb}{\mathbf{b}}
\newcommand{\bhb}{\mathbf{\hat{b}}}
\newcommand{\bB}{\mathbf{B}}
\newcommand{\bC}{\mathbf{C}}
\newcommand{\bc}{\mathbf{c}}
\newcommand{\bd}{\mathbf{d}}
\newcommand{\bhd}{\mathbf{\hat{d}}}
\newcommand{\bD}{\mathbf{D}}
\newcommand{\bhD}{\mathbf{\hat{D}}}
\newcommand{\be}{\mathbf{e}}
\newcommand{\bE}{\mathbf{E}}
\newcommand{\bF}{\mathbf{F}}
\newcommand{\bbf}{\mathbf{\bar{f}}}
\newcommand{\bbF}{\mathbf{\bar{F}}}
\newcommand{\mybf}{\mathbf{f}}
\newcommand{\mybtf}{\mathbf{\tilde{f}}}
\newcommand{\bG}{\mathbf{G}}
\newcommand{\bg}{\mathbf{g}}
\newcommand{\bH}{\mathbf{H}}
\newcommand{\bh}{\mathbf{h}}
\newcommand{\tH}{\mathbf{\tilde{H}}}
\newcommand{\bI}{\mathbf{I}}
\newcommand{\bJ}{\mathbf{J}}
\newcommand{\bk}{\mathbf{k}}
\newcommand{\bK}{\mathbf{K}}
\newcommand{\bL}{\mathbf{L}}
\newcommand{\btL}{\mathbf{\tilde{L}}}
\newcommand{\bM}{\mathbf{M}}
\newcommand{\bN}{\mathbf{N}}
\newcommand{\bn}{\mathbf{n}}
\newcommand{\btP}{\mathbf{\tilde{P}}}
\newcommand{\bP}{\mathbf{P}}
\newcommand{\bQ}{\mathbf{Q}}
\newcommand{\bp}{\mathbf{p}}
\newcommand{\bq}{\mathbf{q}}
\newcommand{\btq}{\mathbf{\tilde{q}}}
\newcommand{\bpt}{\mathbf{\tilde{p}}}
\newcommand{\bR}{\mathbf{R}}
\newcommand{\bs}{\mathbf{s}}
\newcommand{\bS}{\mathbf{S}}
\newcommand{\bbs}{\mathbf{\bar{s}}}
\newcommand{\bbS}{\mathbf{\bar{S}}}
\newcommand{\btS}{\mathbf{\tilde{S}}}
\newcommand{\btT}{\mathbf{\tilde{T}}}
\newcommand{\bT}{\mathbf{T}}
\newcommand{\bt}{\mathbf{t}}
\newcommand{\bu}{\mathbf{u}}
\newcommand{\bU}{\mathbf{U}}
\newcommand{\bV}{\mathbf{V}}
\newcommand{\bv}{\mathbf{v}}
\newcommand{\bW}{\mathbf{W}}
\newcommand{\bw}{\mathbf{w}}
\newcommand{\bx}{\mathbf{x}}
\newcommand{\btx}{\mathbf{\tilde{x}}}
\newcommand{\bX}{\mathbf{X}}
\newcommand{\by}{\mathbf{y}}
\newcommand{\bty}{\mathbf{\tilde{y}}}
\newcommand{\bhy}{\mathbf{\hat{y}}}
\newcommand{\bby}{\mathbf{\bar{y}}}
\newcommand{\bY}{\mathbf{Y}}
\newcommand{\bz}{\mathbf{z}}
\newcommand{\bZ}{\mathbf{Z}}
\newcommand{\btz}{\mathbf{\tilde{z}}}
\newcommand{\bhz}{\mathbf{\hat{z}}}
\newcommand{\bl}{\mathbf{l}}
\newcommand{\cR}{\mathcal{R}}
\newcommand{\btheta}{\mathbf{\theta}}
\newcommand{\bff}{\mathbf{f}}
\newcommand{\gausstwo}[2]{\mathcal{N}(#1; #2)}

\newcommand{\gauss}[3]{\ensuremath{\mathcal{N}(#1 | #2 ; #3)}}

\newcommand{\barqj}{{\bar{q}_j}}

\newcommand{\pctab}{\hspace{0.2in}}
\newenvironment{pseudocode} {\begin{center} \begin{minipage}{\textwidth}
                             \normalsize \vspace{-2\baselineskip} \begin{tabbing}
                             \pctab \= \pctab \= \pctab \= \pctab \=
                             \pctab \= \pctab \= \pctab \= \pctab \= \\}
                            {\end{tabbing} \vspace{-2\baselineskip}
                             \end{minipage} \end{center}}
\newenvironment{items}      {\begin{list}{$\bullet$}
                              {\setlength{\partopsep}{\parskip}
                                \setlength{\parsep}{\parskip}
                                \setlength{\topsep}{0pt}
                                \setlength{\itemsep}{0pt}
                                \settowidth{\labelwidth}{$\bullet$}
                                \setlength{\labelsep}{1ex}
                                \setlength{\leftmargin}{\labelwidth}
                                \addtolength{\leftmargin}{\labelsep}
                                }
                              }
                            {\end{list}}
\newcommand{\newfun}[3]{\noindent\vspace{0pt}\fbox{\begin{minipage}{3.3truein}\vspace{#1}~ {#3}~\vspace{12pt}\end{minipage}}\vspace{#2}}

\newcommand{\key}{\textbf}
\newcommand{\fun}{\textsc}

\maketitle

\begin{abstract}
Scene parsing has attracted a lot of attention in computer vision. While parametric models have proven effective for this task, they cannot easily incorporate new training data. By contrast, nonparametric approaches, which bypass any learning phase and directly transfer the labels from the training data to the query images, can readily exploit new labeled samples as they become available. Unfortunately, because of the computational cost of their label transfer procedures, state-of-the-art nonparametric methods typically filter out most training images to only keep a few  relevant ones to label the query. As such, these methods throw away many images that still contain valuable information and generally obtain an unbalanced set of labeled samples. In this paper, we introduce a nonparametric approach to scene parsing that follows a sample-and-filter strategy. More specifically, we propose to sample labeled superpixels according to an image similarity score, which allows us to obtain a balanced set of samples. We then formulate label transfer as an efficient filtering procedure, which lets us exploit more labeled samples than existing techniques. Our experiments evidence the benefits of our approach over state-of-the-art nonparametric methods on two benchmark datasets.
\end{abstract}


\begin{figure}[t!]

\includegraphics[width=0.22\textwidth,height=90pt]{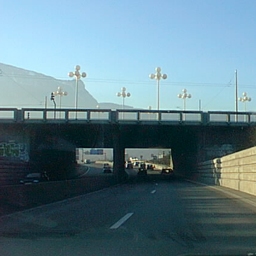}
\hspace*{\fill} 
\includegraphics[width=0.22\textwidth,height=90pt]{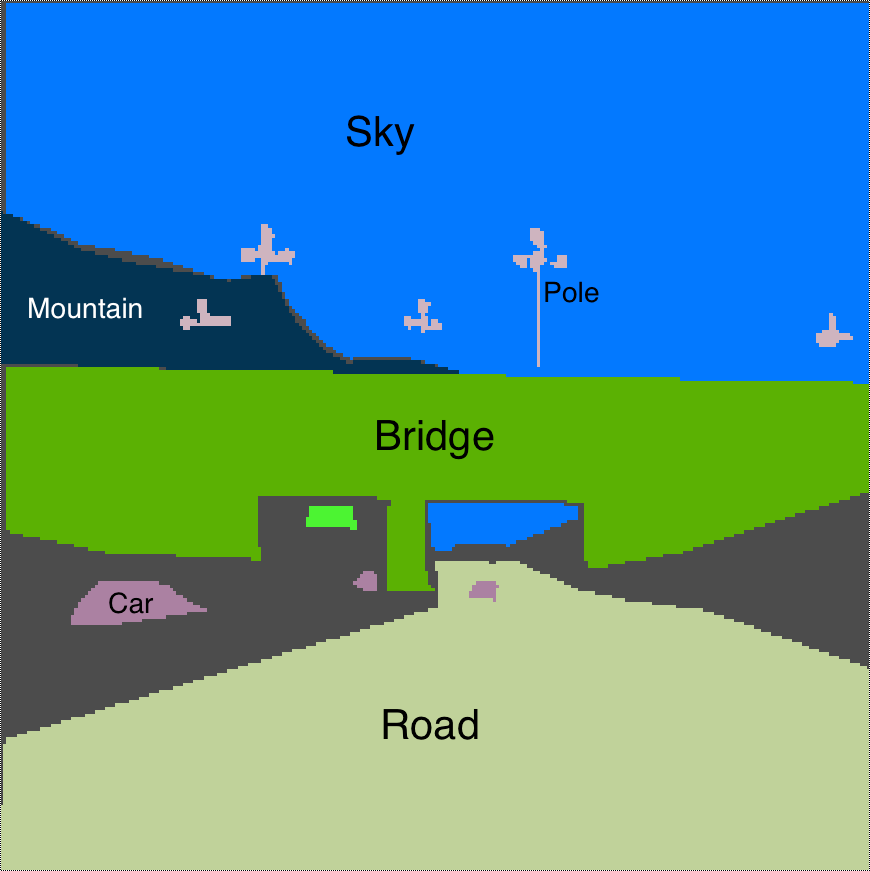}

\includegraphics[width=0.22\textwidth,height=90pt]{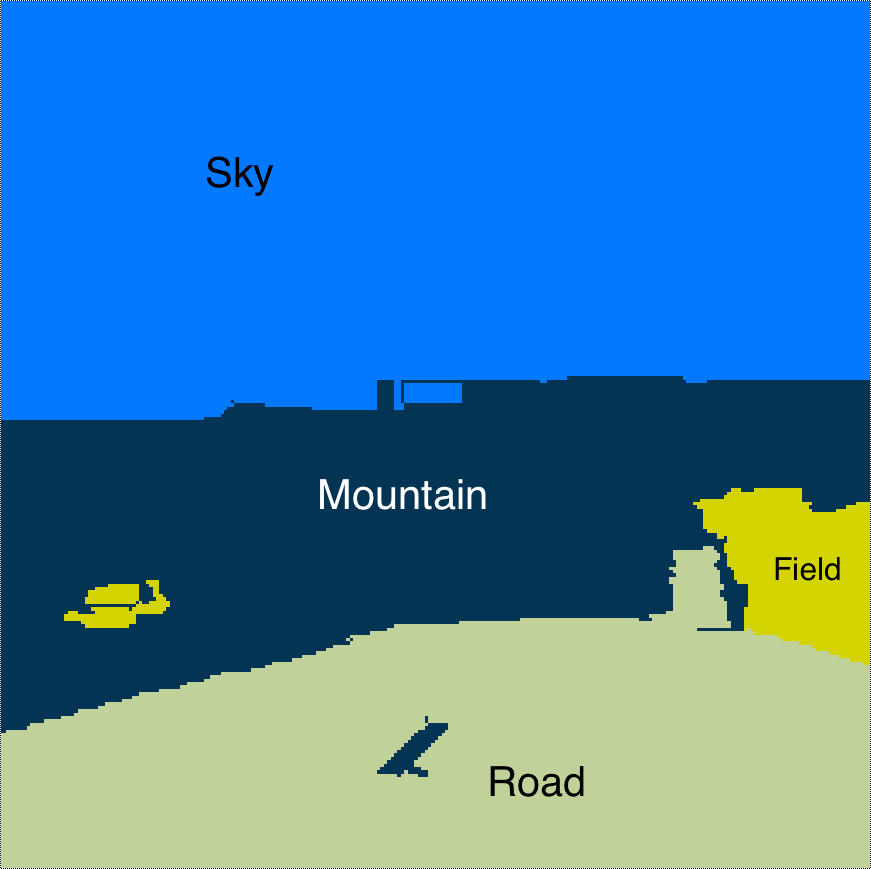}
\hspace*{\fill} 
\includegraphics[width=0.22\textwidth,height=90pt]{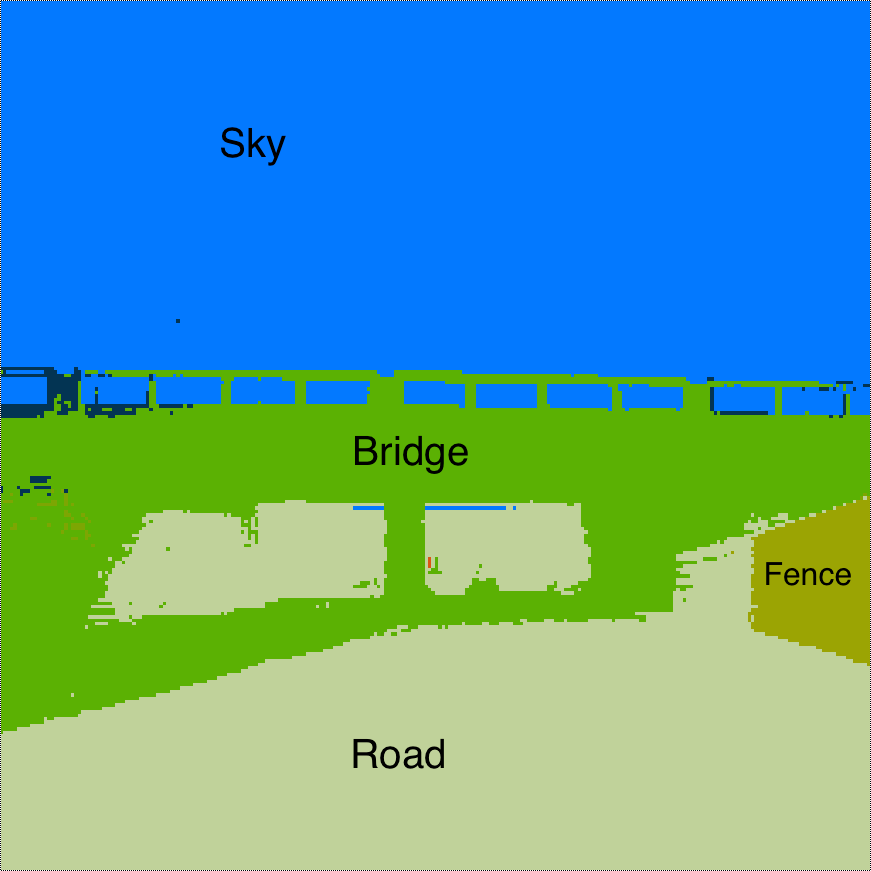}

\caption{{\bf Top left:} Query image; {\bf Top right:} ground truth; {\bf bottom left:} Superparsing method~\cite{Tighe:ijcv:2013}; {\bf bottom right:} Our method.} \label{fig:titlepage}
\end{figure}

\section{Introduction}

Scene parsing, also known as semantic segmentation, tackles the problem of assigning one class label to every pixel in an image (Fig.~\ref{fig:titlepage}). The traditional approach to addressing this problem consists of having a separate training phase that learns a parametric model, which will then be applied to the test data~\cite{Shotton:textonboost:eccv, Ladicky:pami:2013, Kohli:2009:RHO, Gould:ICCV09, Ladicky:IJCV:2013, Kraehenbuehl:icml:2013, Liu:CVPR2015, Tighe:Finding:2015, Yang:rare:2015, Marian:Widerange:2015, Farabet:CNL:2013, Sharma:DeepParsing:2015, long:fcn:2015, Shuai:paramAndNonparam:2015}. While effective, this approach doesn't account for the dynamic nature of our world, where images are constantly being acquired. Indeed, as new training data becomes available, these techniques need to re-train their model. Unfortunately, this process is generally very time-consuming; for example, training a state-of-the-art Convolutional Neural Network (CNN) can take several days.

Nonparametric methods have recently emerged as a solution to this drawback~\cite{Liu:pami:2011, Eigen:cvpr:2012, Gould:eccv:2012, Tighe:ijcv:2013, Myeong:tensor:2013, Myeong:context:2012, Singh:cvpr:2013, Tung:Collage:eccv}. Rather than training a model, these techniques aim at directly transferring the semantics of labeled images to the test data. As such, they can readily incorporate new labeled images as they become available.


Most nonparametric methods~\cite{Liu:pami:2011, Eigen:cvpr:2012, Tighe:ijcv:2013, Myeong:tensor:2013, Myeong:context:2012, Singh:cvpr:2013, Tung:Collage:eccv} follow a two-stage procedure: They first retrieve a set of images similar to the query image, and then transfer the labels of these retrieved images to the query.  The retrieval step plays two important roles. First, it discards the labeled images that are irrelevant to the query. Second, by reducing the amount of data to take into account, it effectively speeds up the transfer step. While the benefits of the former point are unquestionable, the latter one is somewhat more dubious and mostly motivated by the relative lack of speed of the transfer step. Indeed, to remain fast, existing techniques typically throw away images which might still contain valuable information. This particularly causes problems when the classes are unbalanced, since the less-frequent classes might easily not even appear in the retrieved images.

In this paper, we introduce an approach to scene parsing that follows a sample-and-filter strategy. 
Specifically, instead of retrieving a fixed number of similar training images, we randomly sample the labeled superpixels from the training data according to an image-similarity score. We then formulate label transfer as a Gaussian filtering procedure, which computes the label of a query superpixel from the labels of sampled superpixels. Thanks to the efficiency of our filtering procedure and to our sampling strategy, our approach lets us (i) make use of more labeled superpixels than existing retrieval-based techniques; and (ii) obtain a set of labeled samples that is more balanced in terms of class frequency.

We evaluate our method on two large-scale benchmark datasets, SIFTFlow~\cite{Liu:pami:2011} and LM-SUN~\cite{Tighe:ijcv:2013}. Our experiments evidence the benefits of our approach in terms of both accuracy and computation time over state-of-the-art nonparametric scene parsing techniques.

\section{Related Work}

In recent years, scene parsing has attracted a lot of attention. In particular, many methods have proposed to tackle scene parsing by first learning a model from training data, and then applying this model to the unseen test data. A popular trend among these methods consists of learning a pixel classifier and use it as a unary potential in a Markov Random Field (MRF), which models the dependencies of the class labels of two or more pixels~\cite{Shotton:textonboost:eccv, Ladicky:pami:2013, Kohli:2009:RHO, Gould:ICCV09, Ladicky:IJCV:2013, Kraehenbuehl:icml:2013, Liu:CVPR2015}. When it comes to the classifier itself, several directions have been proposed, such as boosting-based classifiers~\cite{Shotton:textonboost:eccv, Yang:rare:2015, Marian:Widerange:2015}, or exemplar-based object detectors~\cite{Tighe:Finding:2015, Liu:CVPR2015}. With the recent advent of deep learning, several works have focused on developing CNNS to perform semantic segmentation~\cite{Farabet:CNL:2013, Sharma:DeepParsing:2015, long:fcn:2015, Shuai:paramAndNonparam:2015}. While effective, these approaches are parametric, and thus cannot incorporate new labeled data without a computationally expensive re-training procedure.

By contrast, nonparametric approaches do not learn any model, but instead transfer the labels of the training data to the query images. As a consequence, they can directly incorporate new labeled data. To the best of our knowledge, this idea was first introduced by Liu \etal~\cite{Liu:pami:2011}, who made use of SIFTFlow~\cite{Liu:siftflow:2011} to transfer the labels from a small set of retrieved images to the query. Unfortunately, the computational cost of SIFTFlow significantly affected the speed of their approach. Instead, in~\cite{Gould:eccv:2012}, Gould \& Zhang built on the efficient PatchMatch algorithm~\cite{Barnes:2009:PAR,Barnes:2010:TGP}, which allowed them to bypass the retrieval step and build a graph over the entire training set to perform label transfer. For the algorithm to remain tractable, however, the degree of the vertices in the graph had to be kept low, which, in turn, affected the labeling accuracy.

\emph{Superparsing}, introduced by Tighe \& Lazebnik~\cite{Tighe:ijcv:2013}, probably constitutes the most popular nonparametric approach to scene parsing. From a set of retrieved images, it produces a label for each query superpixel by combining the results of nearest-neighbor retrieval using multiple superpixel features in a na\"{\i}ve Bayes classifier. Inspired by \cite{Tighe:ijcv:2013}, Eigen \& Fergus~\cite{Eigen:cvpr:2012} and Singh \& Kosecka~\cite{Singh:cvpr:2013} proposed to learn weights for the different superpixel features; Myeong \etal~\cite{Myeong:context:2012,Myeong:tensor:2013} incorporated pairwise and higher-order contextual relationships among the object categories into the Superparsing framework; Tung \& Little~\cite{Tung:Collage:eccv} proposed to reason at the level of complete objects, obtained by an objectness criterion, instead of relying on superpixels. While all these modification of Superparsing have indeed led to higher segmentation accuracy, they also come at a higher computational cost. Furthermore, and more importantly, all these methods, including Superparsing, make an initial strong decision to reject a large number of labeled images, many of which might still contain valuable information for the query.

By contrast, here, we introduce a sampling strategy to collect the relevant labeled superpixels, which lets us retrieve a balanced number of samples for each class. Thanks to this sampling procedure, and to our efficient filtering approach to label transfer, our algorithm yields accuracies that are competitive with the state-of-the-art methods, while being significantly faster.

\section{Method}
We now introduce our nonparametric approach to scene parsing. To this end, let $\mathcal{X}'=\{\bx'_1,\bx'_2,\dots,\bx'_{N_t}\}$ denote the set of feature vectors $\bx'_j$ representing the training superpixels, with corresponding ground-truth labels $\mathbf{Y}'=\{y'_1,y'_2,\dots,y'_{N_t}\}$, $y_i \in \{1,\dots,L\}$. Our goal is to transfer these labels to a set of query superpixels encoded by their feature vectors $\mathbf{X}=\{\bx_1,\bx_2,\dots,\bx_{N_q}\}$. As mentioned above, here, we follow a sample-and-filter approach, which first randomly samples a balanced set of relevant training superpixels, and then performs label transfer via efficient Gaussian filtering. In the remainder of this section, we present these two steps in detail.

\subsection{Sampling Balanced Superpixels} \label{sec:sampling}

It is undeniable that, as suggested by other nonparametric approaches~\cite{Tighe:ijcv:2013,Eigen:cvpr:2012,Singh:cvpr:2013,Tung:Collage:eccv}, many images from the training data are irrelevant to label the query image. Following this intuition and common practice, we therefore first rank the training images according to their similarity to the query image using the method explained in Section~\ref{sec:ranking}. At this stage, state-of-the-art nonparametric scene parsing algorithms~\cite{Tighe:ijcv:2013,Eigen:cvpr:2012,Singh:cvpr:2013,Tung:Collage:eccv} simply discard the images beyond a pre-defined rank. This, however, typically discards many images with relevant information because of noise in the ranking process and because the pre-defined rank is usually chosen so as to keep few images. Furthermore, with this process, the number of retrieved superpixels belonging to each class is typically unbalanced.

By contrast, here, we propose to make use of the ranks to randomly sample training superpixels. To this end, we assign a dissimilarity value 
\begin{equation}
d_j \in \left\{\frac{1}{N_t},\frac{1}{N_t-1},\dots,1\right\}
\label{eq:dissimilarity}
\end{equation}
to each training superpixel according to the rank of its corresponding image. Note that the superpixels in the image with the highest rank, \ie, the image most similar to the query, will be assigned the lowest dissimilarity value. From these dissimilarity values, we compute a score for each superpixel as 
\begin{equation}
p_j = \exp\left(-\frac{d_j^2}{\sigma_d}\right)\;,\;\;\forall\; j \in \{1,2\dots,N_t\}\;.
\end{equation}
We then use this score to randomly sample the superpixels using the method proposed in \cite{Wong:1980:datasample}. Ultimately, while superpixels with larger values $p_j$ are more likely to be picked, this still potentially allows any superpixel to be selected.

Furthermore, and more importantly, since we randomly sample superpixels, and each superpixel is assigned a class label, we can enforce having a balanced set of training data by sampling the same number of superpixels for each class. Note that, in practice, this is not always possible, since some classes truly occur very rarely in the training data. This will be addressed in the label transfer step of our approach. Nevertheless, our sampling procedure produces a more balanced set of superpixels than the simple image retrieval strategy. Furthermore, thanks to our efficient filtering approach to label transfer, discussed below, we can exploit more labeled superpixels than state-of-the-art nonparametric scene parsing techniques.

\subsubsection{Image Ranking} \label{sec:ranking}
As mentioned above, our sampling strategy relies on an image ranking procedure that reflects the similarity between each training image and the query. This procedure works as follows. 
We extract three  global image descriptors, \ie, spatial pyramid of color histograms, GIST \cite{Oliva:2001:GIST} and Histogram of Oriented Gradients (HOG) visual words~\cite{sunDatabase:cvpr:2010}, from each image in the training set and from the query. We then produce three rankings according to the similarity of each of these descriptors, using the $\chi^2$ distance metric. 
The final rank of the images are then obtained by sorting their average ranks over these three rankings.

\subsection{Label Transfer via Efficient Filtering}
The sampling procedure of Section~\ref{sec:sampling} produces a balanced set of $N_s$ training superpixels encoded by feature vectors $\{\bx'_1,\bx'_2,\dots,\bx'_{N_s}\}$. Our goal now is to transfer the labels of these superpixels to those of the query image. Here, we propose to formulate label transfer as an efficient Gaussian filtering operation.

To this end, let $\mathbf{q}'_j$ be the $L$-dimensional binary vector encoding the label of the $j^{\rm th}$ training superpixel as
\begin{equation}
   \mathbf{q}'_j(l) = \begin{cases}
               1               & y'_j=l\\
               0			 & \text{otherwise}\;,
           \end{cases}
\label{qVector}
\end{equation}
where $\bq'_j(l)$ indicates the $l^{\rm th}$ element of $\bq'_j$. We then propose to estimate the label of the query superpixels as
\begin{equation}
\begin{aligned}
\mathbf{q}_i = \sum_{j=1}^{N_s}k(\mathbf{x}_i,\mathbf{x}'_j)\mathbf{q}'_j
\;, \;\;
\forall \; i\in \{1,2,\dots,N_q\}
\end{aligned}
\label{filtering}
\end{equation}
where $k(\mathbf{x_i},\mathbf{x'_j})$ is a Gaussian kernel that encodes how similar two superpixels are in terms of their feature vectors $\mathbf{x_i}$ and $\mathbf{x'_j}$, and thus how strongly we believe that these two superpixels should have the same label. The specific form of kernel used in our experiments is given in Section~\ref{sec:kernelFeatures}.

\begin{figure*}
\begin{subfigure}{0.31\textwidth}
\includegraphics[width=\linewidth]{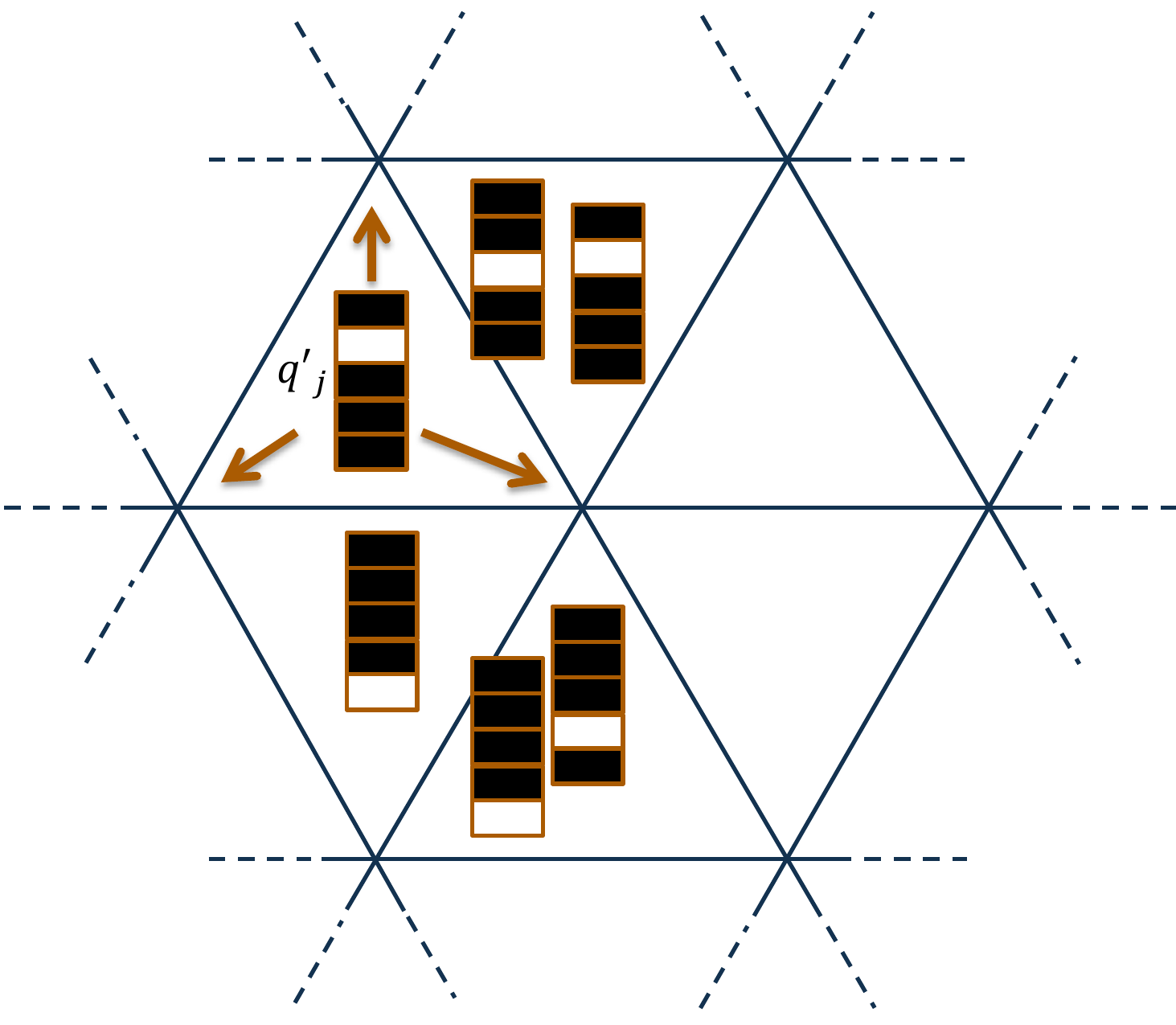}
\caption{Splatting} \label{fig:1a}
\end{subfigure}
\hspace*{\fill} 
\begin{subfigure}{0.31\textwidth}
\includegraphics[width=\linewidth]{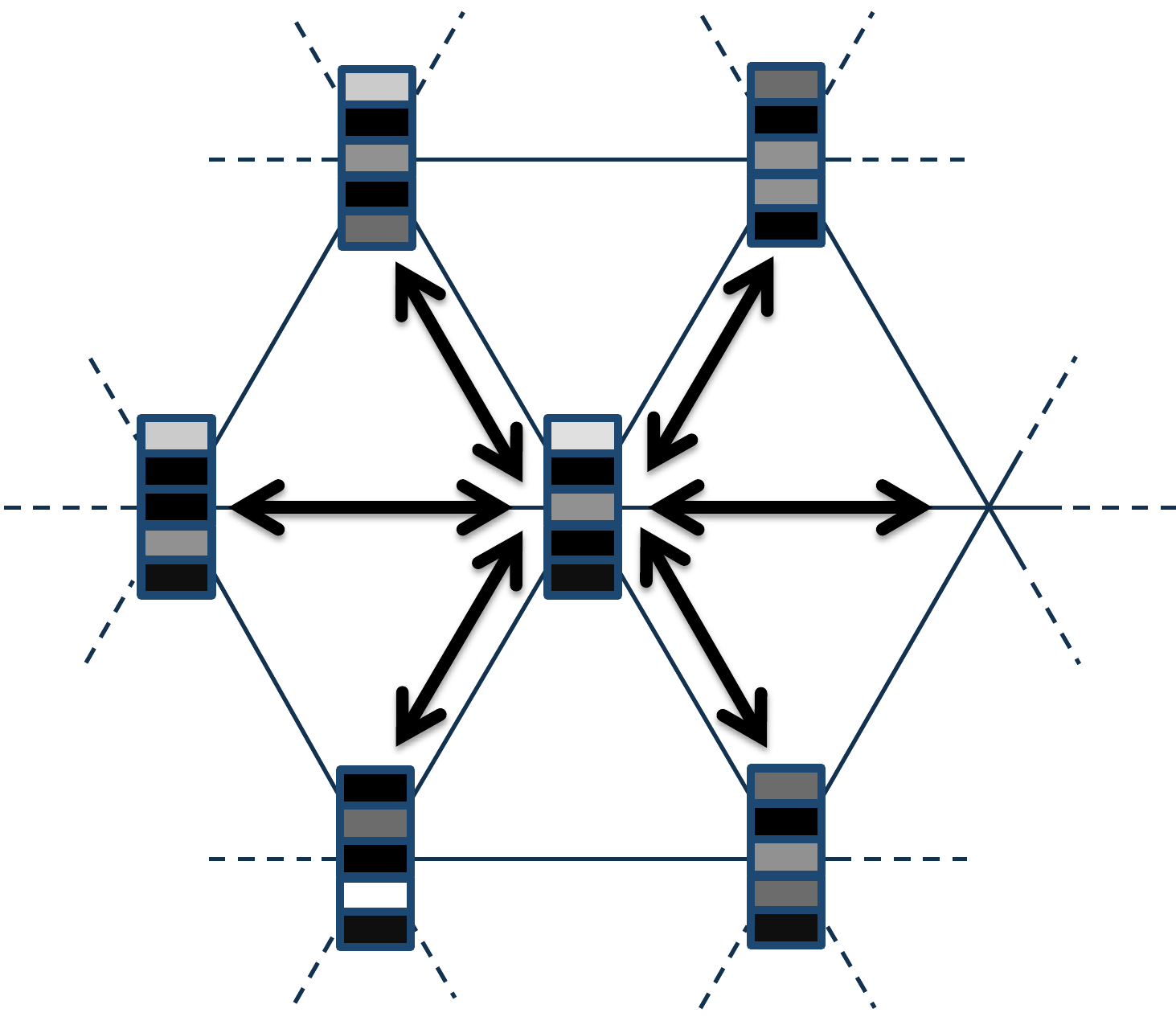}
\caption{Blurring} \label{fig:1b}
\end{subfigure}
\hspace*{\fill} 
\begin{subfigure}{0.31\textwidth}
\includegraphics[width=\linewidth]{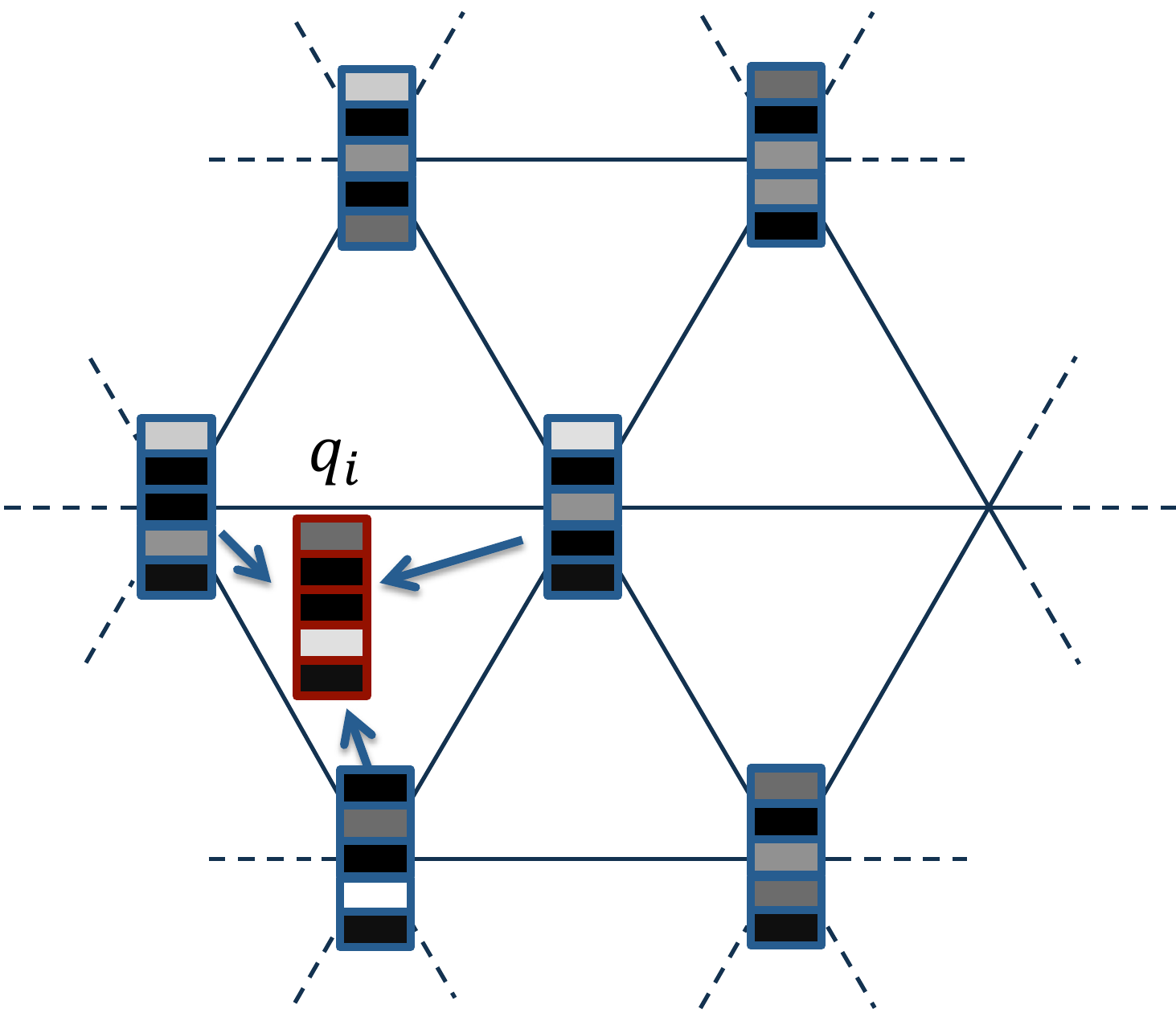}
\caption{Slicing} \label{fig:1c}
\end{subfigure}
\caption{A schematic of the filtering process on a permutohedral lattice. The block structures represent label vectors, and the gray-level intensities in each vector denote the likelihoods of different classes. Fig.~\ref{fig:1a} shows how the binary label vector $\mathbf{q'_j}$ (defined in Eq.~\ref{qVector}) is mapped onto the lattice vertices. The blurring step is depicted in Fig.~\ref{fig:1b} where Gaussian blurring is applied to the lattice points. Fig.~\ref{fig:1c} illustrates the slicing step where a query data receives label information from the lattice vertices. } \label{fig:permutohedral}
\end{figure*}

Since Eq.~\ref{filtering} involves $N_s$ summations for every query superpixel, the total computational complexity for a query image would be $O(N_sN_q)$. For large numbers of retrieved superpixels, which is what we advocate here, this approach would thus be prohibitively costly. However, Eq.~\ref{filtering} corresponds to a Gaussian filtering operation, for which fast and accurate approximations have been proposed~\cite{Adams:permutohedral:2010,Gastal:2011:DTE,Adams:2009:GKF}. In particular, here, we make use of the permutohedral lattice-based formulation of~\cite{Adams:permutohedral:2010}. This method relies on three steps, illustrated in Fig.~\ref{fig:permutohedral}: {\it Splatting}, which, in our case, consists of mapping the training data to the permutohedral lattice and computing the values at the vertices of the lattice; {\it Blurring}, which approximates the Gaussian filter locally, thus updating the values at the vertices of the lattice; and {\it Slicing}, which, in our case, consists of mapping the query superpixels to the lattice and computing their values as a linear combination of the values of a few vertices. The first two steps, which only involve the training data, can be performed in $O(N_s)$. For each query superpixel, slicing can be done in constant time, \ie, linearly dependent on the feature dimension, but not on $N_s$. Altogether, this therefore yields a total computational complexity of $O(N_s+N_q)$. 


\subsubsection{Kernels}\label{sec:kernelFeatures}
In this work, we define the kernel of Eq.~\ref{filtering} as 
\begin{equation}
k(\bx_i,\bx'_j) = w_1k_1(\bx_i,\bx'_j) + w_2k_2(\bx_i,\bx'_j)\;,
\label{multipleKernels}
\end{equation}
where $k_1$ and $k_2$ are two Gaussian kernels defined below. Note that the algorithm described above translates easily to the two-kernel case by simply making use of two permutohedral lattices, and, for each query superpixel, combining the two predicted label vectors.

In practice, as a first kernel, we make use of a color-based Gaussian, expressed as
\begin{equation}
\begin{aligned}
k_1(\bx_i,\bx'_j) = \textrm{exp}\Bigg(-\frac{\|\bc_i-\bc'_j\|^2}{{\sigma^2}_c}-\frac{|t_i-t'_j|^2}{{\sigma^2}_t}\\
-\frac{|s_i-s'_j|^2}{{\sigma^2}_{s}}-\frac{|d_i-d'_j|^2}{{\sigma^2}_d}\Bigg)\;,
\end{aligned}
\label{kernel1}
\end{equation}
where $\bc$ is the vector of average RGB intensities of a superpixel, $s$ is the standard deviation of the gray-level intensities in the superpixel, $t$ is the minimum distance of the superpixel to the top of the image, and $d$ is the dissimilarity value defined in Eq.~\ref{eq:dissimilarity}. Note that we set $d_i=0$ for the query superpixels. 

The second kernel relies on the image gradient and is defined as
\begin{equation}
\begin{aligned}
k_2(\bx_i,\bx'_j) = \textrm{exp}\Bigg(-\frac{\|\bh_i-\bh'_j\|^2}{{\sigma^2}_h}-\frac{|t_i-t'_j|^2}{{\sigma^2}_t}\\
-\frac{|s_i-s'_j|^2}{{\sigma^2}_{s}}-\frac{|d_i-d'_j|^2}{{\sigma^2}_d}\Bigg)\;,
\end{aligned}
\label{kernel2}
\end{equation}
where $\bh$ is the 6-bin HOG descriptor of the superpixels. 

In our experiments, the standard deviations $\sigma_{c}$, $\sigma_{t}$, $\sigma_{s}$, $\sigma_{d}$ and $\sigma_{h}$, and the weights $w_1$ and $w_2$ were obtained using a validation set. 

\subsubsection{Handling Rare Classes}
As mentioned in Section~\ref{sec:sampling}, while we aim at selecting a balanced set of training superpixels, having exactly an equal number for each class is not always possible, due to the insufficient number of superpixels in some {\it rare} classes. As a matter of fact, this problem occurs frequently in large-scale datasets, and would have a negative impact on the filtering procedure. Indeed, in Eq.~\ref{filtering}, the contribution of a superpixel belonging to a rare class and highly similar to the query superpixel could easily be dominated by the combined contributions of superpixels from a common class, even if they are not too similar to the query.

To address this problem, we propose to modify the definition of $\mathbf{q}'_j$ in Eq.~\ref{filtering} as
\begin{equation}
   \mathbf{q}'_j(l) = \begin{cases}
               \lambda(l)               & y'_j=l\\
               0			 & \text{otherwise}\;,
           \end{cases}
\label{qVectorScaled}
\end{equation}
where $\lambda(l) = N_{max}/N(l)$, with $N_{max}$ the maximum number of samples picked from any class, and $N(l)$ the number of samples picked from class $l$. 
The term $\lambda$ approaches one for the frequent categories, whereas it increases the contribution of the superpixels belonging to rare classes in the filtering process. Note that, in the perfectly balanced case, all classes have again the same influence.

\begin{algorithm*}
	\KwData{Query image + entire set of training images} 
    \caption{Label Transfer via Efficient Filtering}
    \BlankLine
    {\it Rank the training images based on their similarity to the query image} ~~(Section~\ref{sec:ranking}) \\
    {\it Random sampling on the training superpixels given their similarity scores ($R$)} ~~(Section~\ref{sec:sampling}) \\
    \For{$i = 1$ \KwTo $N_q$}{
        	$\mathbf{q}_i = \sum_{j=1}^{N_t}k(\mathbf{x}_i,\mathbf{x}'_j)\mathbf{q}'_j$ \tcp*{Filtering the training superpixels}
    }
    $\bu_i = -\textrm{log}(\btq_i)$ \tcp*{Compute a unary term based on the normalized filtered labels}
    {\it Compute the pixelwise location prior}  ~~(Section~\ref{sec:crf}) \\
    {\it Solve a dense pixel-wise CRF} ~~(Section~\ref{sec:crf})\\ 
    \Return{Dense pixelwise labeling of the query image}
    
\label{alg:the_alg}
\end{algorithm*} 

\subsection{CRF}\label{sec:crf}
The semantic information transferred to the query superpixels by our approach is of course prone to error. As is commonly done in nonparamatric scene parsing methods~\cite{Tighe:ijcv:2013,Singh:cvpr:2013,Myeong:context:2012,Myeong:tensor:2013,Eigen:cvpr:2012,Tung:Collage:eccv,Gould:eccv:2012}, we therefore make use of a CRF to further smooth these initial predictions. More precisely, our predictions act as unary terms in a CRF defined over the pixels of the query image, which thus prevents us from having to train a classifier. 

Specifically, let $\btq_i$ be the normalized version of the $\bq_i$ obtained from Eq.~\ref{filtering}. We then define the unary potential of each superpixel $i$ as the negative logarithm of $\btq_i$, and assign this unary potential to all the pixels within superpixel $i$. We further combine this unary with a location prior computed as a class histogram built for each pixel from the 15 top images in our ranking. We then make use of the fully-connected CRF model of Kr\"ahenb\"uhl \& Koltun~\cite{Kraehenbuehl:icml:2013}, which relies on an efficient mean-field-based inference strategy to produce a pixelwise labeling of the query image. 


The main steps of our nonparametric scene parsing approach are summarized in Algorithm~\ref{alg:the_alg}.

\section{Experiments}
We evaluated our method on two large-scale datasets, SIFTFlow~\cite{Liu:pami:2011} and LM-SUN~\cite{Tighe:ijcv:2013}. Below, we compare our results with those of state-of-the-art nonparametric scene parsing algorithms.

\subsection{SIFTFlow Dataset}
SIFTFlow~\cite{Liu:pami:2011} consists of 2,688 images taken from outdoor scenes and annotated with 33 different class labels. The standard partition of this dataset includes 2,488 training images and 200 test images. As noted in~\cite{Tighe:ijcv:2013}, this is a difficult dataset due to the large number of rare classes. For this data set, we sampled a maximum of 2500 superpixels of each class. Note, however, that because of rarity, some classes had much fewer samples.

\begin{figure*}

\includegraphics[width=0.15\textwidth]{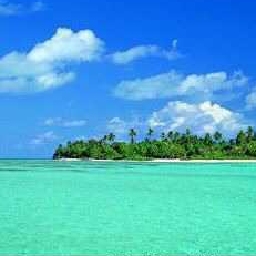}
\hspace*{\fill} 
\includegraphics[width=0.15\textwidth]{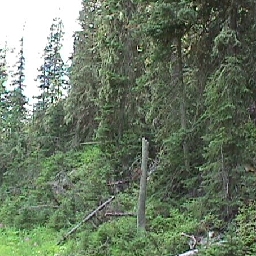}
\hspace*{\fill} 
\includegraphics[width=0.15\textwidth]{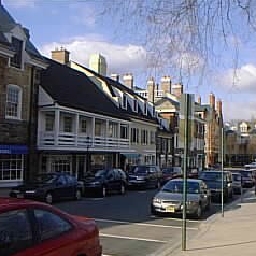}
\hspace*{\fill} 
\includegraphics[width=0.15\textwidth]{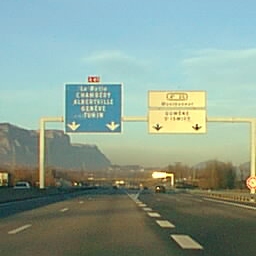}
\hspace*{\fill} 
\includegraphics[width=0.15\textwidth]{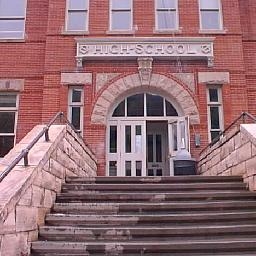}
\hspace*{\fill} 
\includegraphics[width=0.15\textwidth]{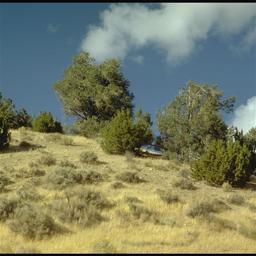}

\includegraphics[width=0.15\textwidth]{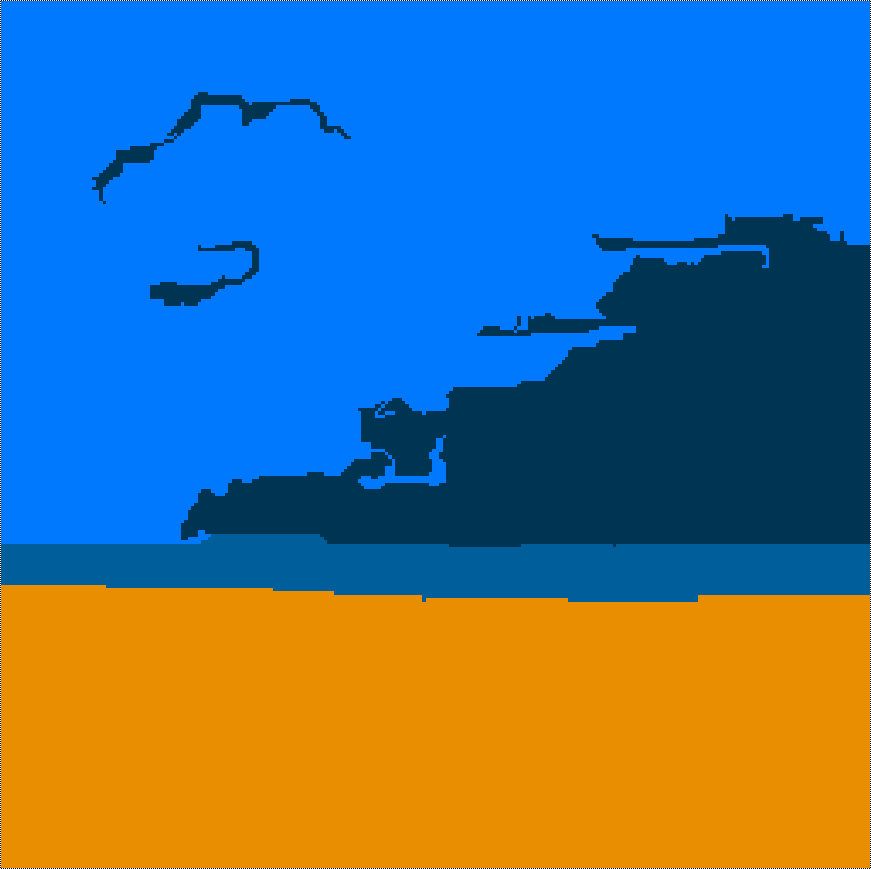}
\hspace*{\fill} 
\includegraphics[width=0.15\textwidth]{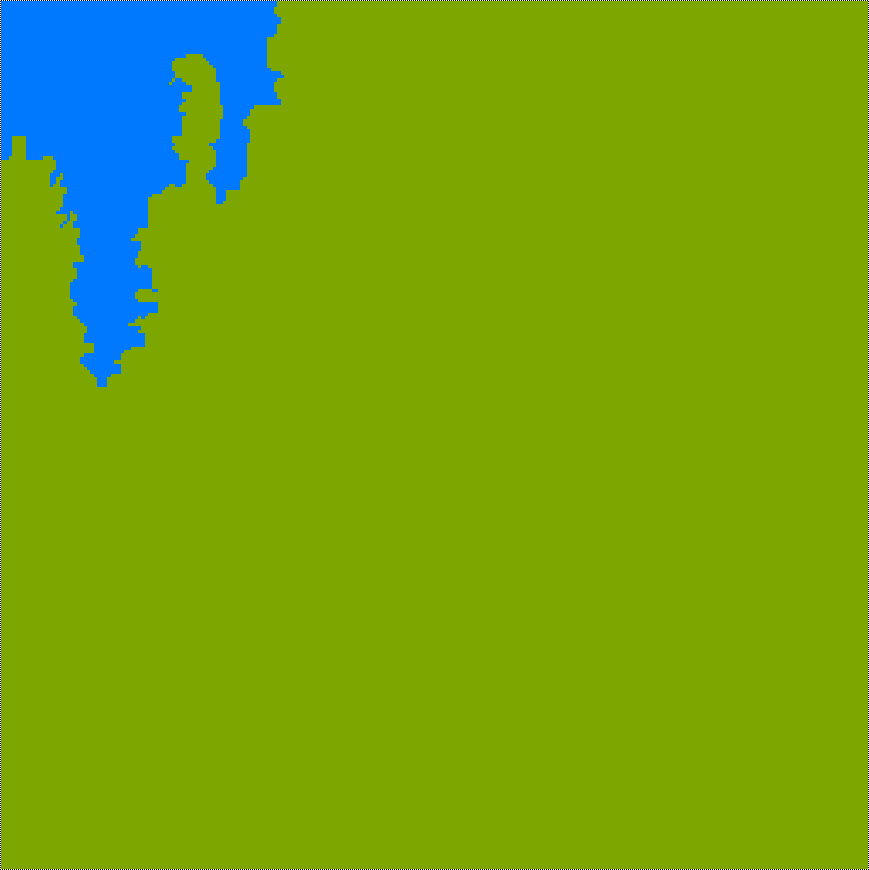}
\hspace*{\fill} 
\includegraphics[width=0.15\textwidth]{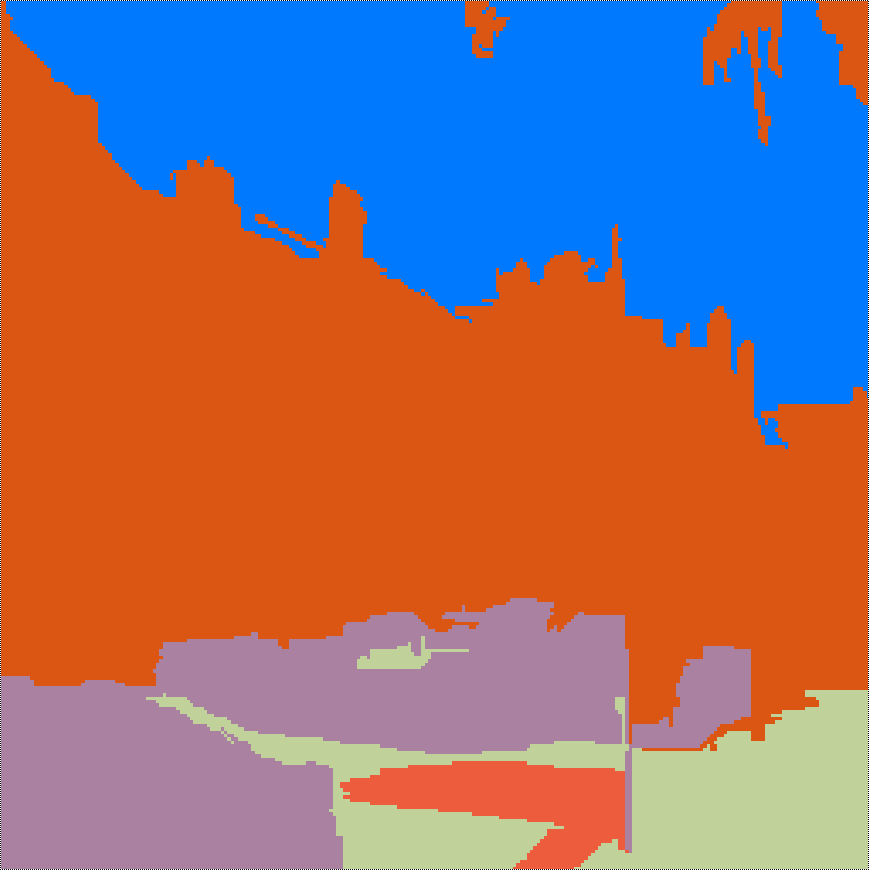}
\hspace*{\fill} 
\includegraphics[width=0.15\textwidth]{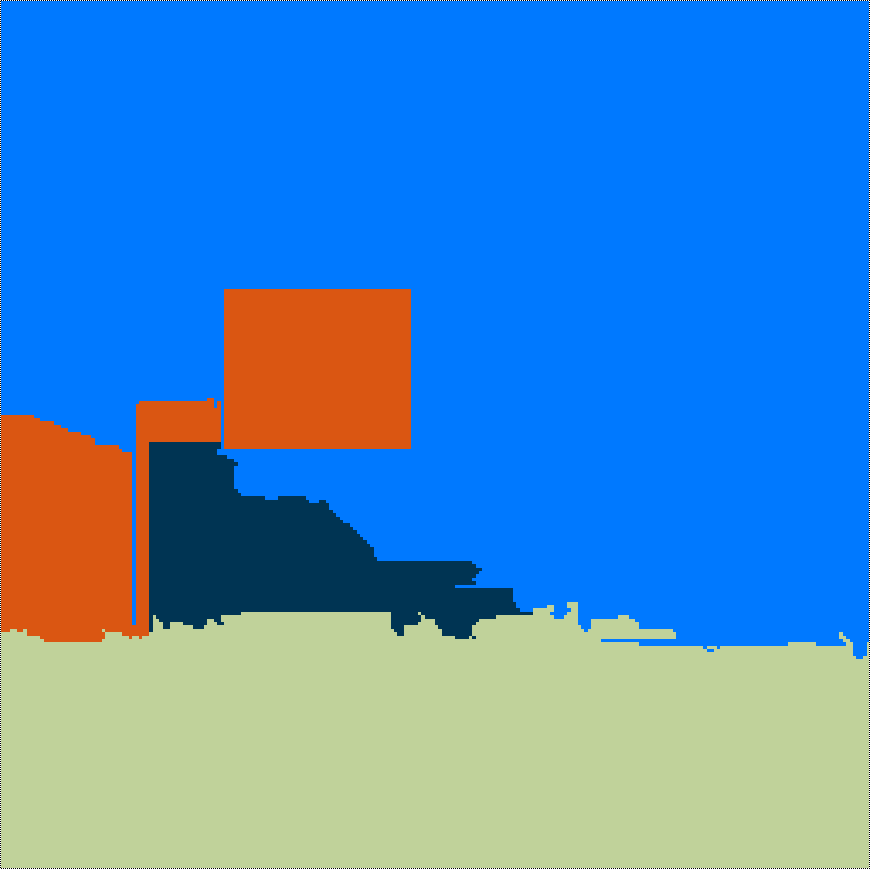}
\hspace*{\fill} 
\includegraphics[width=0.15\textwidth]{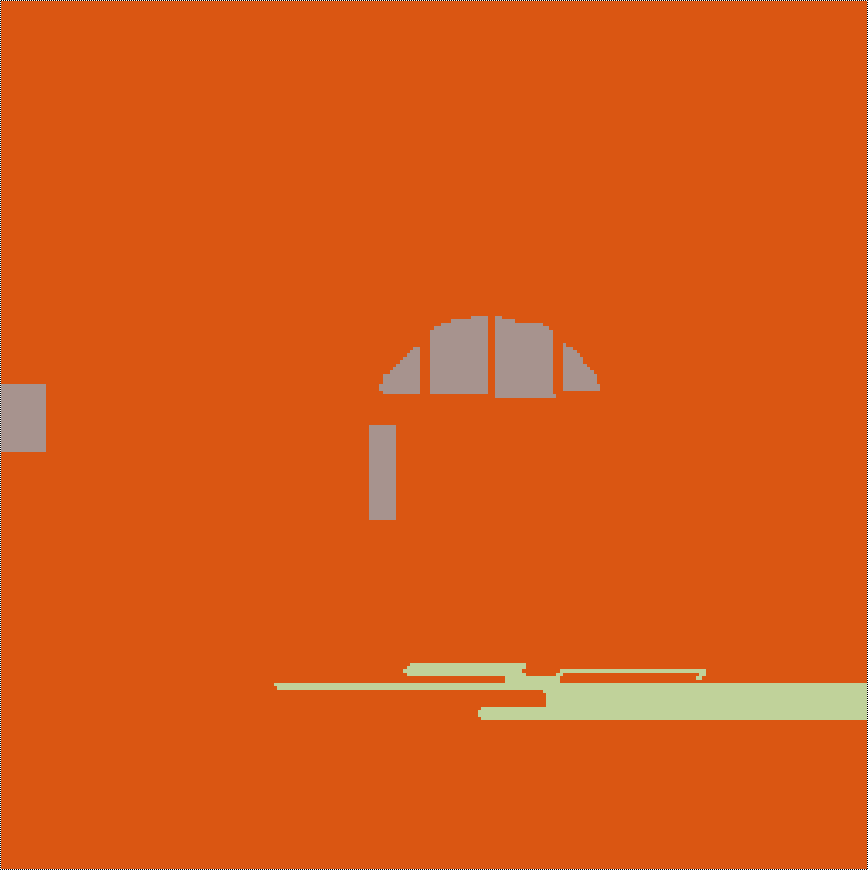}
\hspace*{\fill} 
\includegraphics[width=0.15\textwidth]{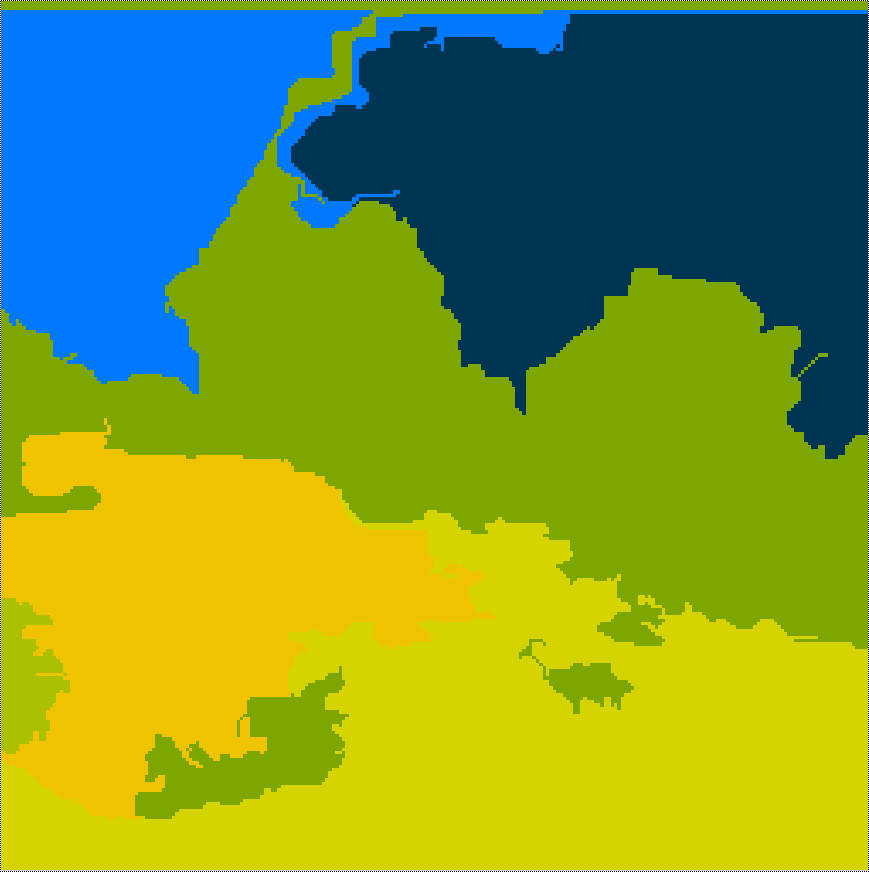}

\includegraphics[width=0.15\textwidth]{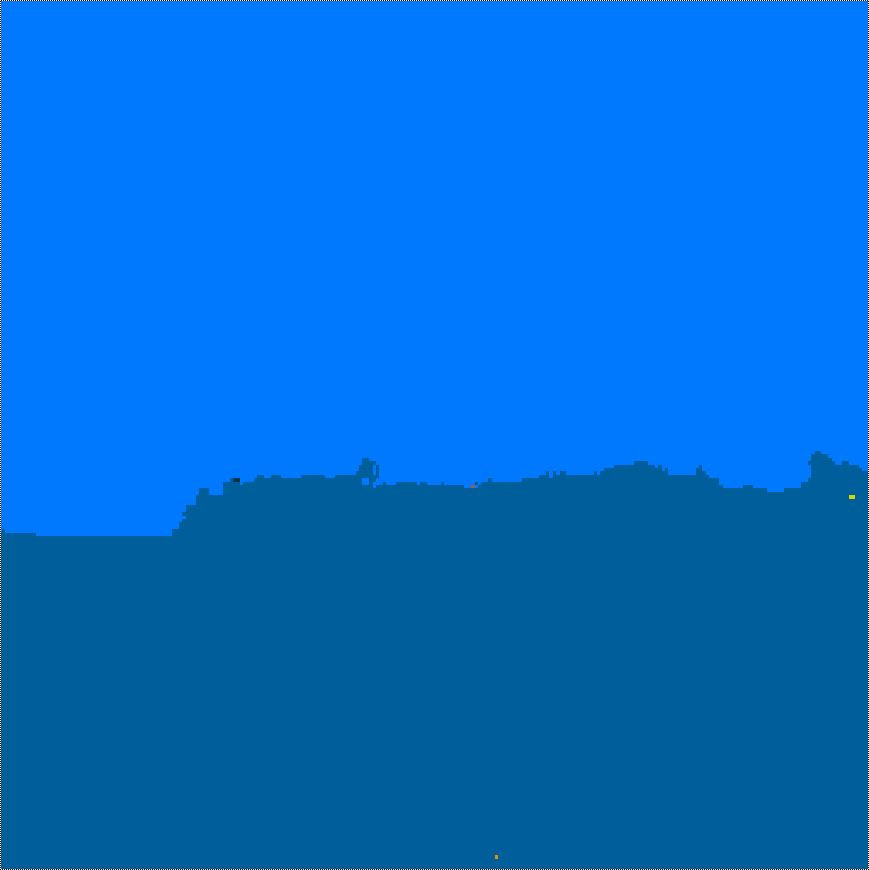}
\hspace*{\fill} 
\includegraphics[width=0.15\textwidth]{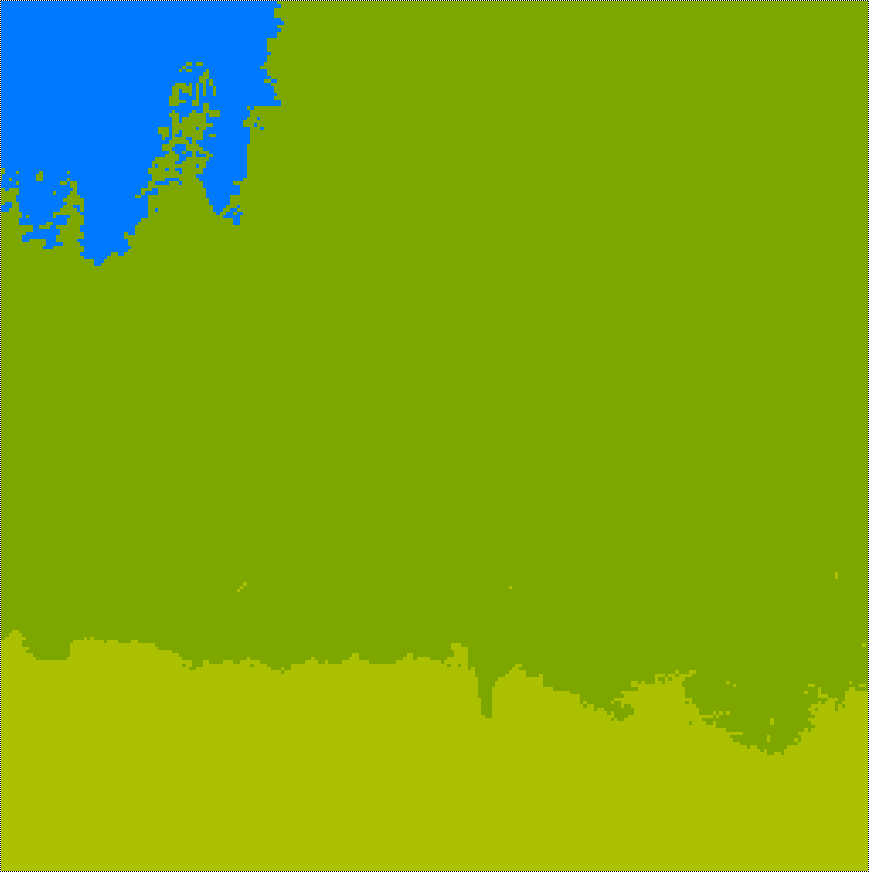}
\hspace*{\fill} 
\includegraphics[width=0.15\textwidth]{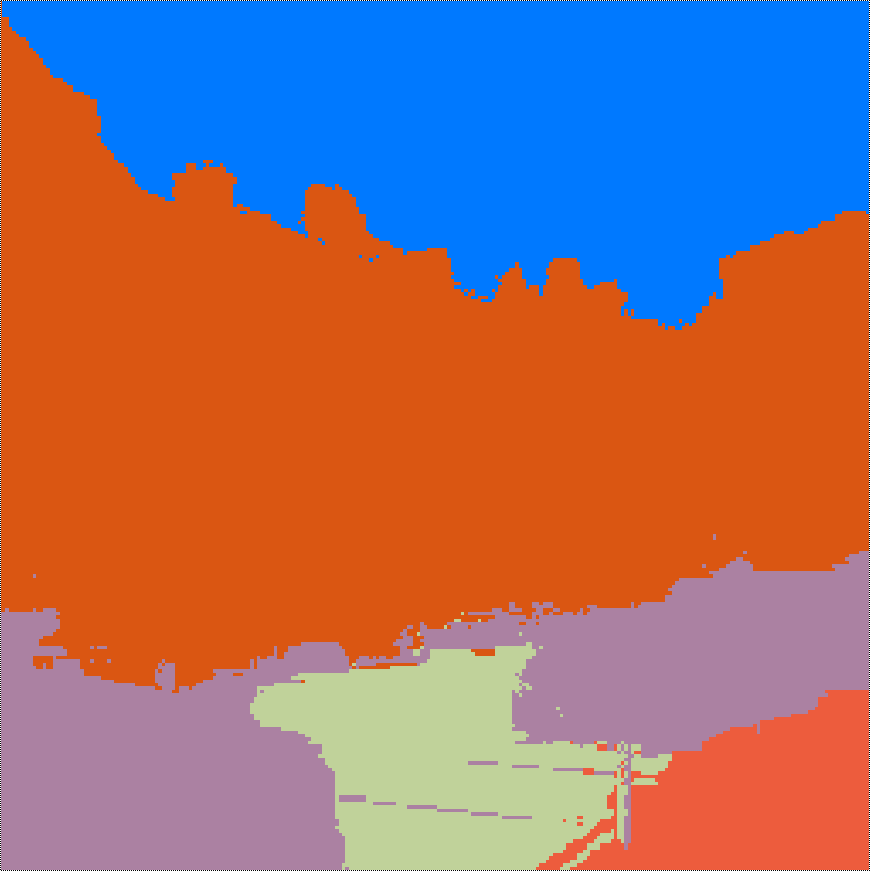}
\hspace*{\fill} 
\includegraphics[width=0.15\textwidth]{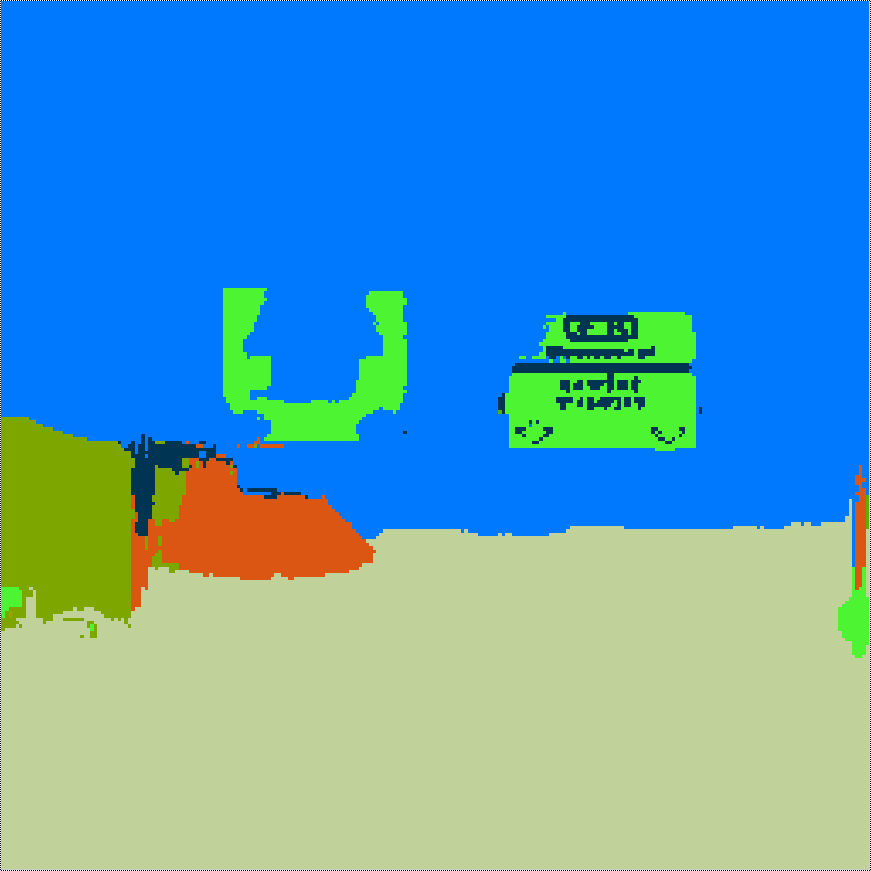}
\hspace*{\fill} 
\includegraphics[width=0.15\textwidth]{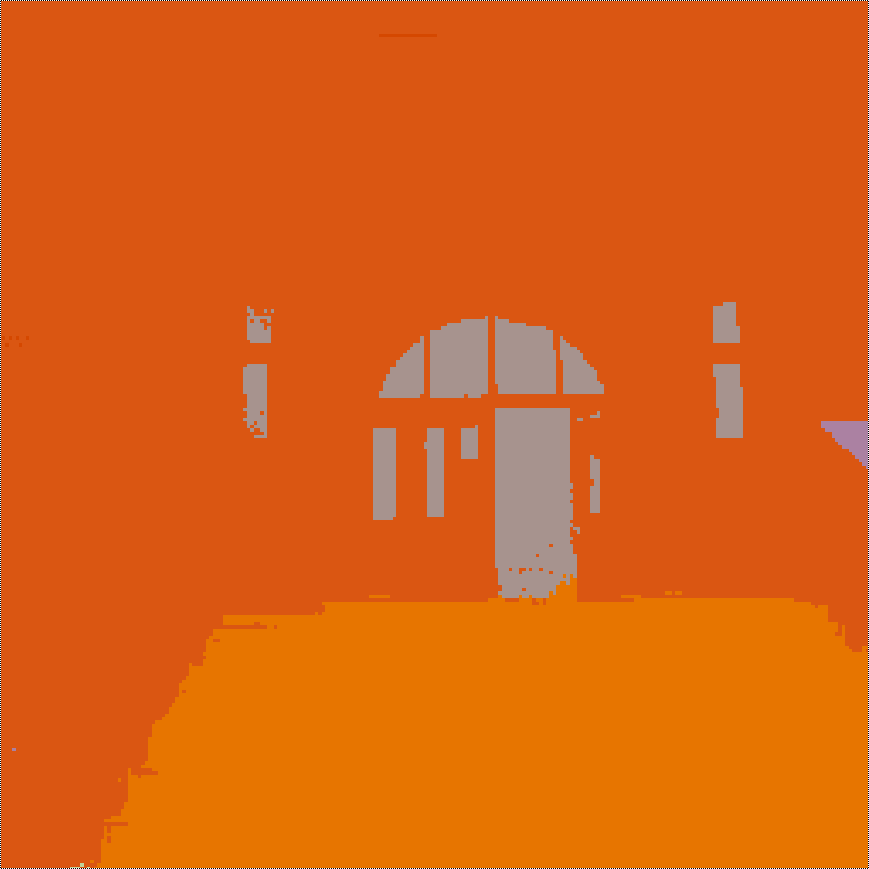}
\hspace*{\fill} 
\includegraphics[width=0.15\textwidth]{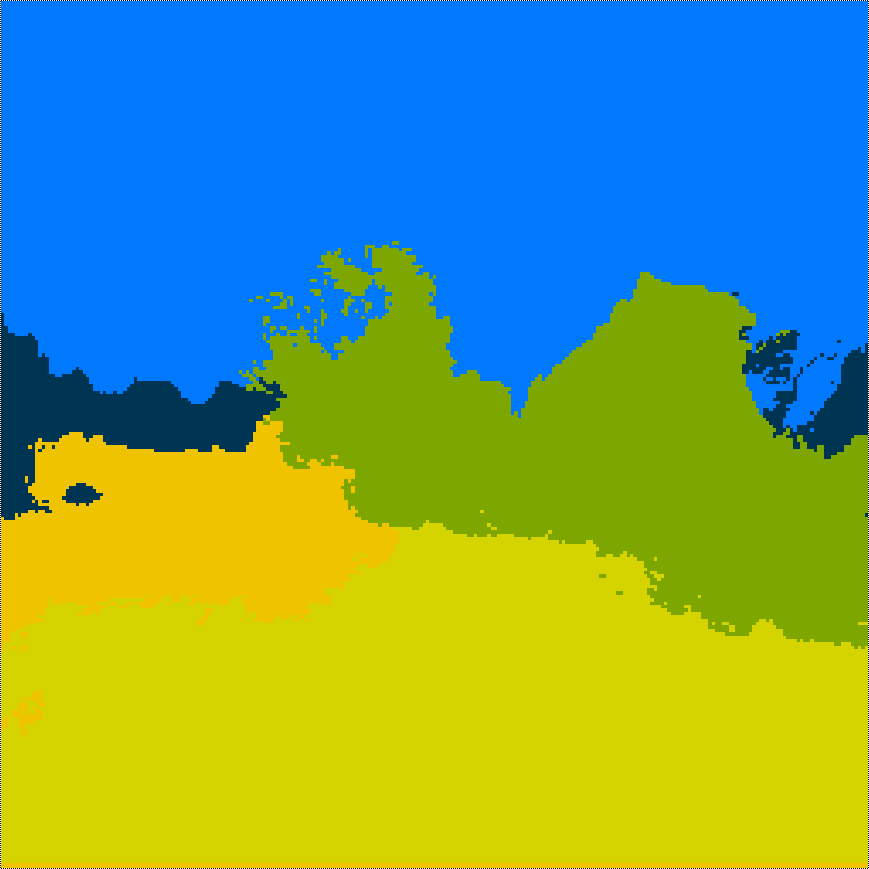}

\includegraphics[width=0.15\textwidth]{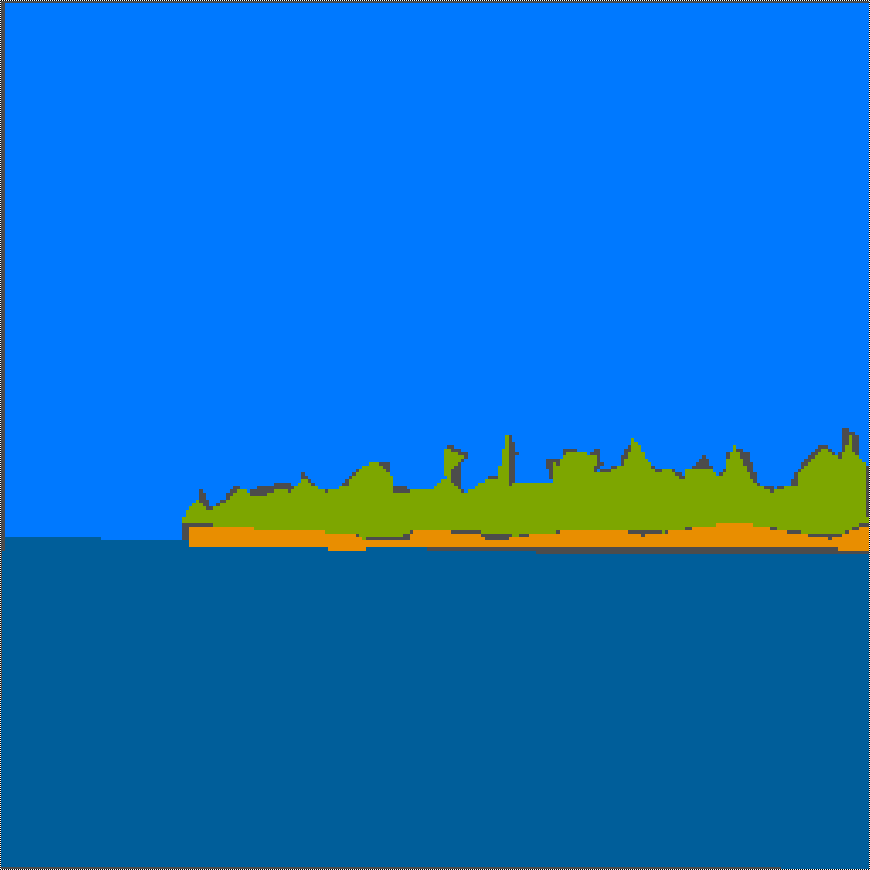}
\hspace*{\fill} 
\includegraphics[width=0.15\textwidth]{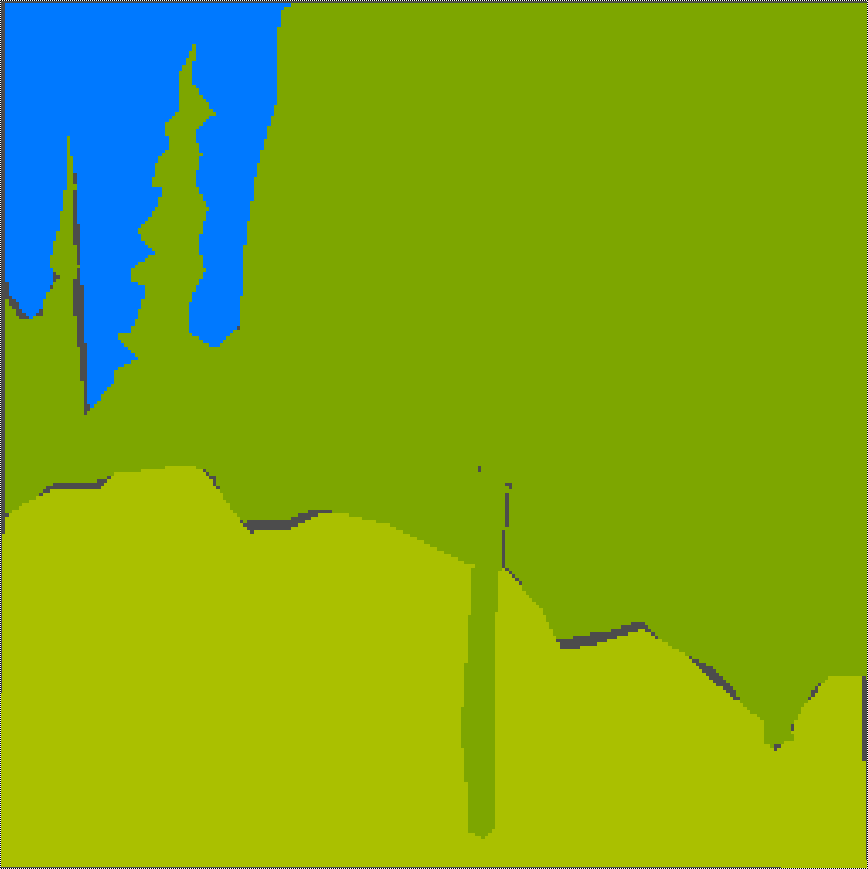}
\hspace*{\fill} 
\includegraphics[width=0.15\textwidth]{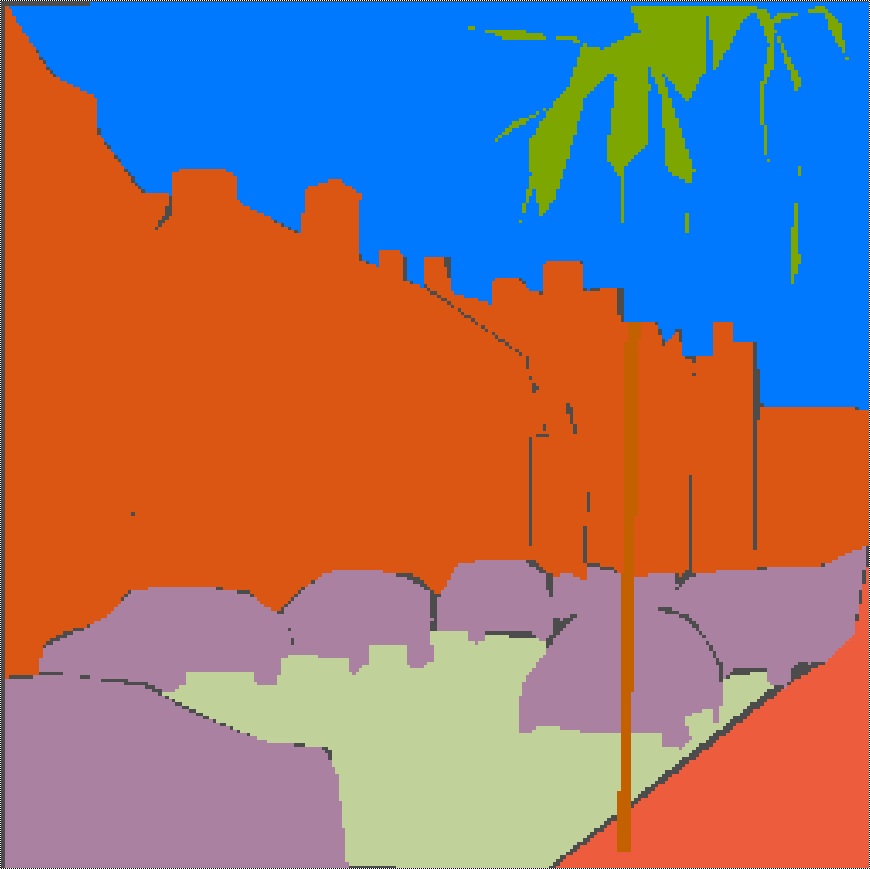}
\hspace*{\fill} 
\includegraphics[width=0.15\textwidth]{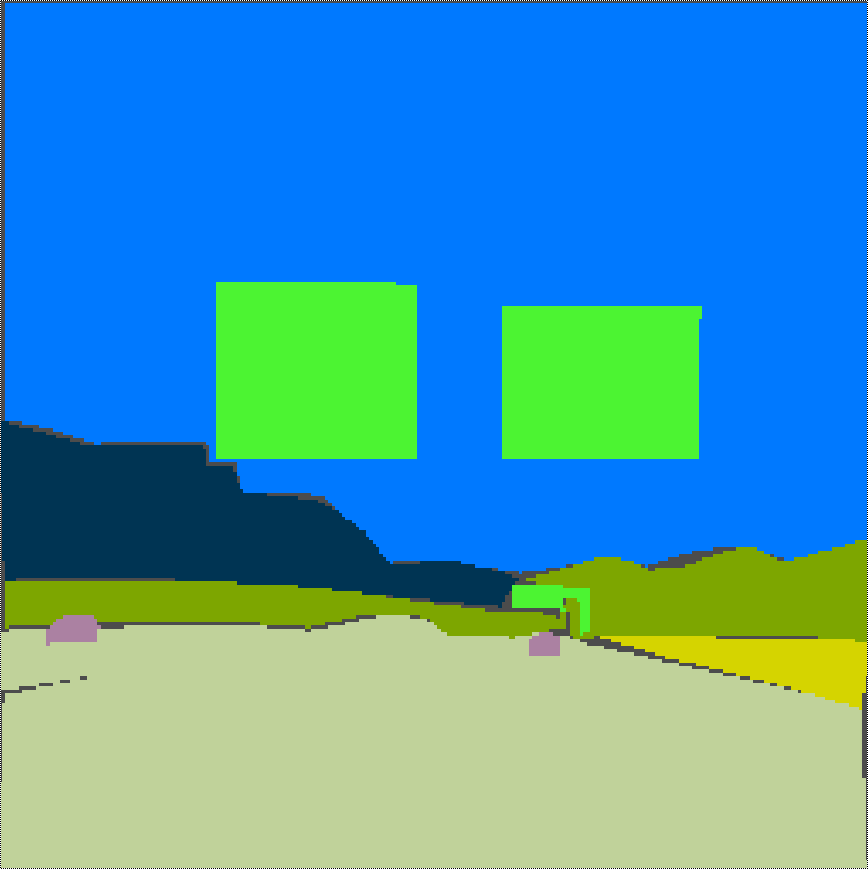}
\hspace*{\fill} 
\includegraphics[width=0.15\textwidth]{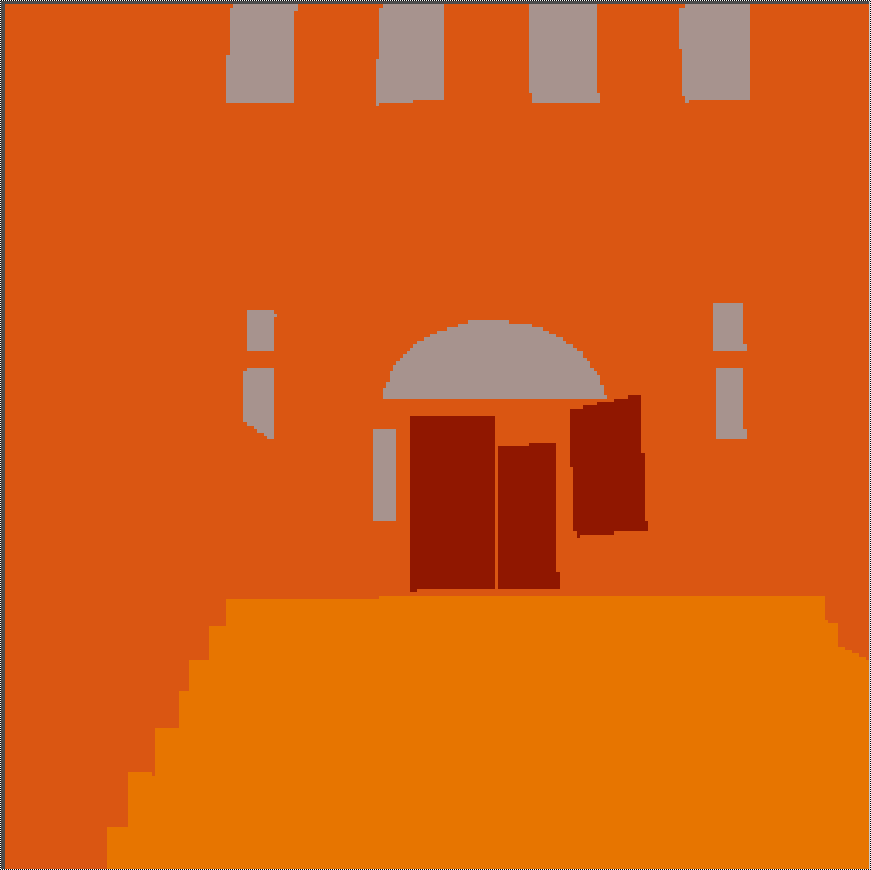}
\hspace*{\fill} 
\includegraphics[width=0.15\textwidth]{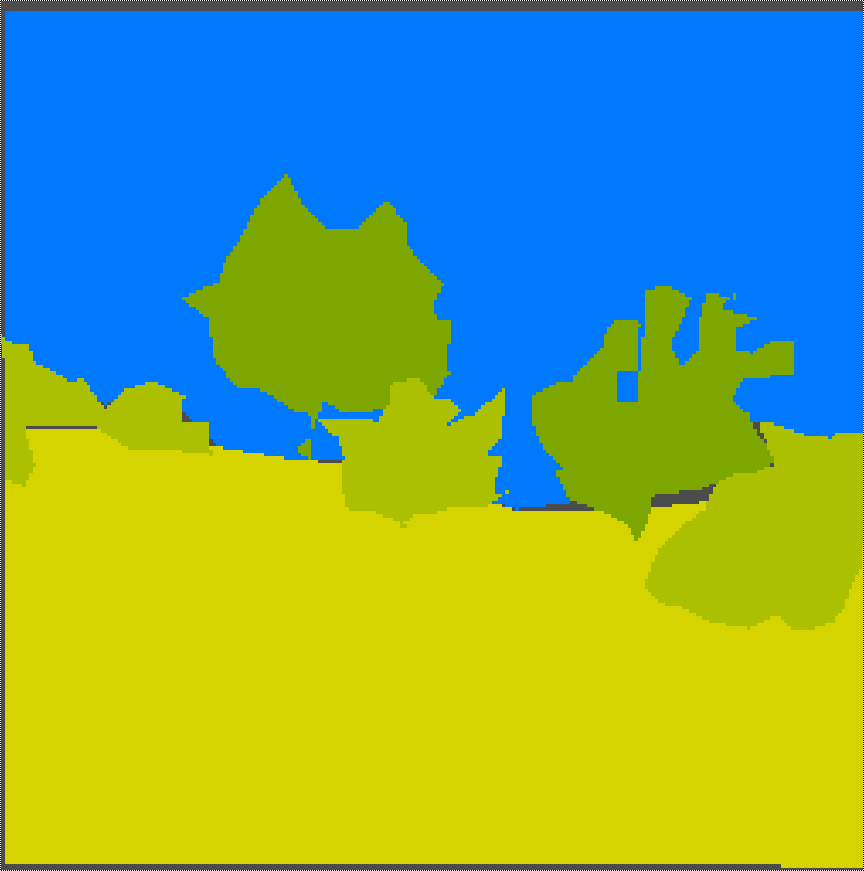}

\includegraphics[width=\textwidth]{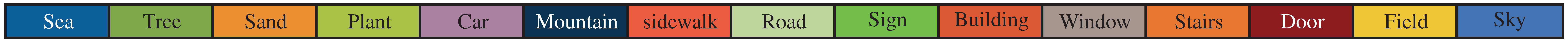}

\caption{Qualitative comparison of our results with those of Superparsing~\cite{Tighe:ijcv:2013} on SIFTFlow. {\bf 1st row:} Query image; {\bf 2nd row:} Superparsing; {\bf 3rd row:} Our approach; {\bf 4th row:} Ground-truth.} \label{fig:siftflow}
\end{figure*}

In Table~\ref{table:siftflowmain}, we compare our results with those of state-of-the-art nonparametric scene parsing methods in terms of per-pixel and average per-class accuracy.
Our approach performs on par with the baselines in per-pixel accuracy, but outperforms most of them in per-class accuracy. This, we believe is due to the more balanced samples that we obtain. To verify this, we replaced our sampling strategy with a fixed retrieval set consisting of all the superpixels of the top 200 images in our ranking.\footnote{We used 200 because it corresponds to the number of images retrieved by the baselines.} Running our filtering-based label transfer procedure on these superpixels resulted in 73.6\% per-pixel accuracy and 22.2\% per-class accuracy. As expected, while the effect on per-pixel accuracy is relatively small, the per-class accuracy decreases dramatically. This clearly evidences the importance of getting as balanced as possible a set of labeled superpixels. Fig.~\ref{fig:siftflow} provides a qualitative comparison of our results with those of Superparsing.

Note that the highest per-class accuracy is achieved by~\cite{Tung:Collage:eccv}. This method, however, relies on an expensive procedure, thus requiring several minutes to process an image. By contrast, thanks to our efficient filtering approach, our algorithm only requires roughly 4 seconds, which outperforms all the baselines.

\begin{table}[t!]\centering\renewcommand{\arraystretch}{1.2}
\caption{Comparison of our approach (Sample \& Filter) with the state-of-the-art nonparametric methods on SIFTFlow.  We report the per-pixel and average per-class accuracies, as well as the average time to process on image. For the baselines, a $>$ indicates that the reported runtimes do not include the entire processing time.}
\vspace{-0.7cm}
\small\addtolength{\tabcolsep}{.01pt}
{
\begin{tabular}{lc|c|c}\\ 
 {}&{\bf \scriptsize per-pixel}&{\bf \scriptsize per-class}&{\bf \scriptsize run-time}\\ 
\hline

{Sample \& Filter}&74.5&35.5&{2.0s}\\

{Sample \& Filter (with CRF)}&76.6&35.0&{4.2s}\\

\hline


{Superparsing (with MRF) ~\cite{Tighe:ijcv:2013}}&{76.2}&{29.1}&{$>$5.9s}\\

{Eigen \etal (with MRF)~\cite{Eigen:cvpr:2012}}&{77.1}&{32.5}&{$>$16.6s}\\

{Myeong \etal (with MRF)~\cite{Myeong:context:2012}}&{77.1}&{32.3}&{$>$23s}\\

{SIFTFlow~\cite{Liu:siftflow:2011}}&{76.7}&{-}&{$>$25mins}\\



{WAKNN (with MRF)~\cite{Singh:cvpr:2013}}&{79.2}&{33.8}&{$>$70s}\\

{CollageParsing (with MRF)~\cite{Tung:Collage:eccv}}&{77.1}&41.1&{2mins}\\

\end{tabular}
}

\label{table:siftflowmain}
\end{table}

\begin{table}[t!]\centering\renewcommand{\arraystretch}{1.2}
\caption{Comparison of our approach (Sample \& Filter) with Superparsing using an ideal image ranking on SIFTFlow.}
\vspace{-0.7cm}
\def\d{.1cm}
\def\r{30}
\small\addtolength{\tabcolsep}{6pt}
{
\begin{tabular}{lc|c}\\ 
 {}&{\bf \scriptsize per-pixel}&{\bf \scriptsize per-class}\\ \hline

{Sample \& Filter (with CRF)}&83.1&44.3\\
\hline

{Superparsing (with MRF)~\cite{Tighe:ijcv:2013}}&{80.2}&{33.6}\\

\end{tabular}
}

\label{table:siftflowideal}
\end{table}

Note that our runtimes were obtained on a standard desktop with an Intel 3.07GHz six-core processor and 12 GB RAM. Our algorithm was implemented mostly in Matlab, with the exception of the filtering step, which was built upon the C++ code of~\cite{Kraehenbuehl:icml:2013}. This leaves a lot of room for speed improvement. While we do not know the exact setup of the baselines, we believe that, since we used an ordinary platform, the runtime comparison remains fair.


To further evaluate the potential of our approach, and following the analysis performed in~\cite{Tighe:ijcv:2013}, we performed an additional experiment based on an {\it ideal} image ranking strategy. To this end, and following~\cite{Tighe:ijcv:2013}, the retrieval was achieved using histograms of ground-truth class labels, both for the training and test images. The idea here is to try and evaluate the best possible performance of our approach. The results of this experiment are reported in Table~\ref{table:siftflowideal}, where we compare our approach with the results of~\cite{Tighe:ijcv:2013} obtained in the same ideal setting. These results indicate that, given a better image similarity measure, our method has the potential to achieve higher accuracy than Superparsing, especially in terms of per-class accuracy.



\begin{figure*}[t!]

\includegraphics[width=0.10\textwidth]{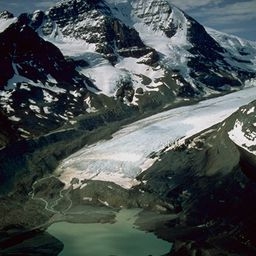}
\hspace*{\fill} 
\includegraphics[width=0.10\textwidth]{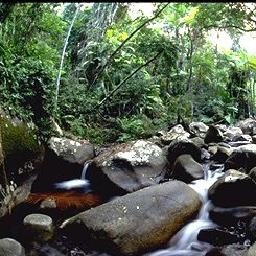}
\hspace*{\fill} 
\includegraphics[width=0.10\textwidth]{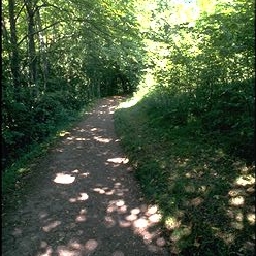}
\hspace*{\fill} 
\includegraphics[width=0.10\textwidth]{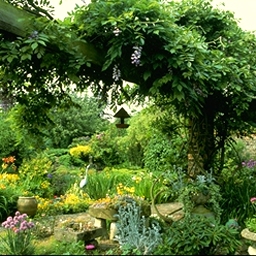}
\hspace*{\fill} 
\includegraphics[width=0.10\textwidth]{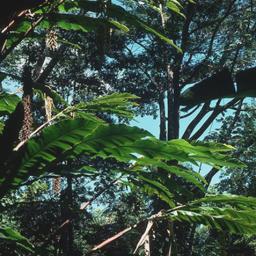}
\hspace*{\fill} 
\includegraphics[width=0.10\textwidth]{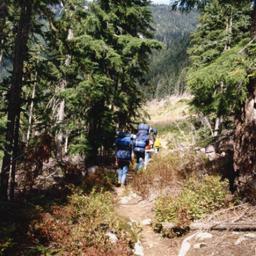}
\hspace*{\fill} 
\includegraphics[width=0.10\textwidth]{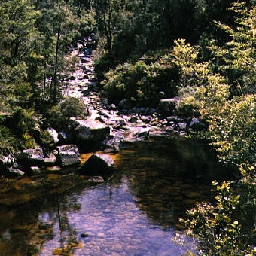}
\hspace*{\fill} 
\includegraphics[width=0.10\textwidth]{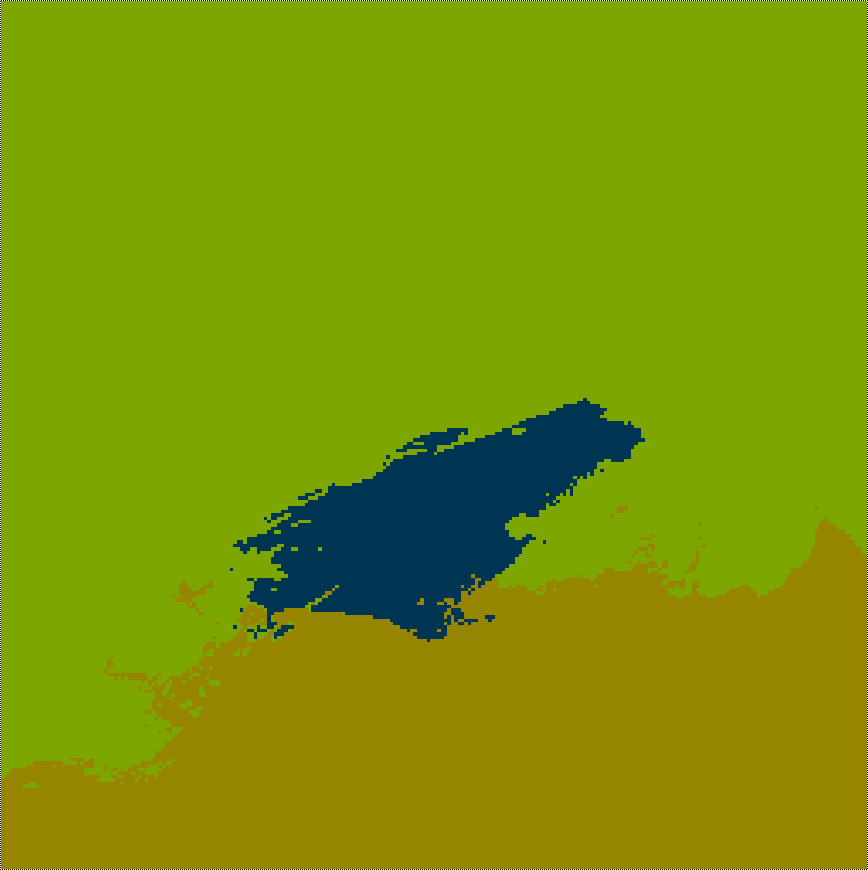}
\hspace*{\fill} 
\includegraphics[width=0.10\textwidth]{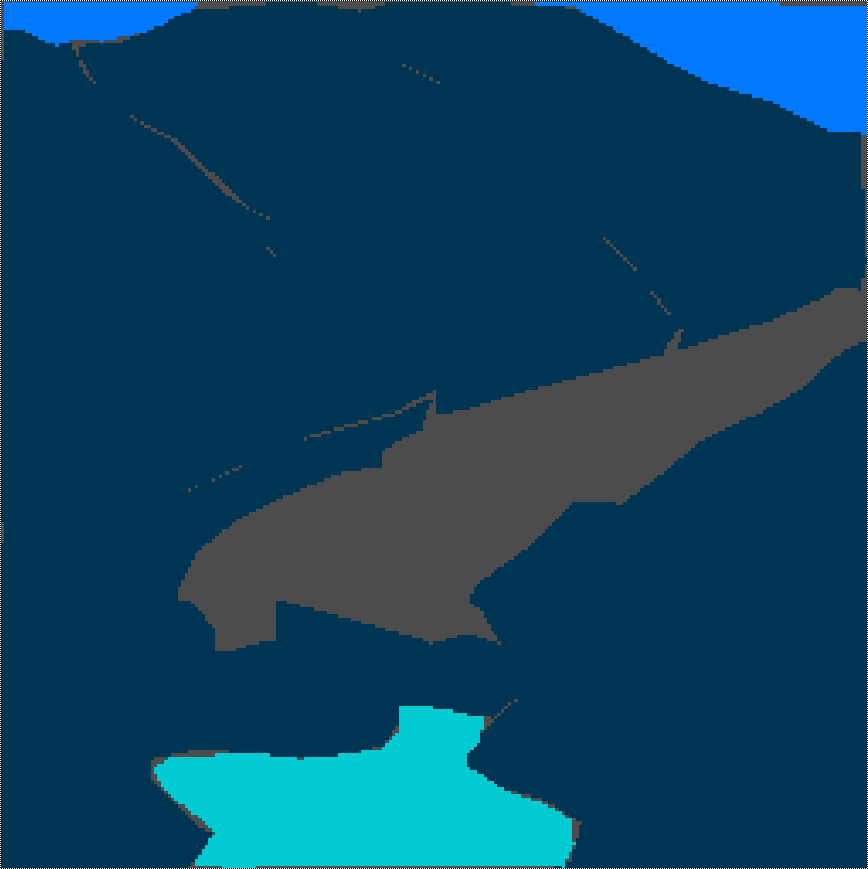}

\caption{{\bf Failure cases.} Poor ranking results yield poor labeling by our approach. The first image is the query and the next six images are top ranked training images. The two last images denote our results and the ground-truth, respectively. Note that the top images in the ranking are semantically irrelevant.} \label{fig:failure}
\end{figure*}

Fig.~\ref{fig:failure} shows a failure case of our method. This figure depicts a query image followed by the top six images in the similarity ranking, the result of our algorithm and the ground-truth. In this case, the image ranking strategy retrieved a semantically irrelevant group of images. As suggested by Table~\ref{table:siftflowideal}, improving the image similarity metric would address this problem. 

\subsection{LM-SUN Dataset}
The LM-Sun dataset~\cite{Tighe:ijcv:2013} is one the most challenging benchmarks available for scene parsing. It includes 45,676 images, among which, following the standard partition, 500 images are taken as test data. The ground-truth annotations of this dataset is comprised of 232 different categories. In this case, we sampled a maximum of 25,000 images per class.

\begin{figure*}

\includegraphics[width=0.15\textwidth,height=100pt]{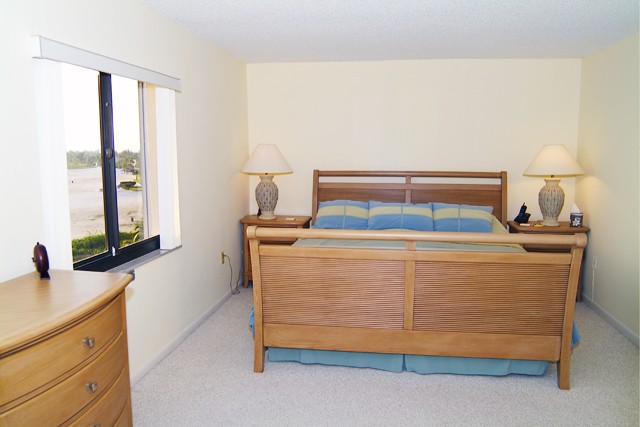}
\hspace*{\fill} 
\includegraphics[width=0.15\textwidth,height=100pt]{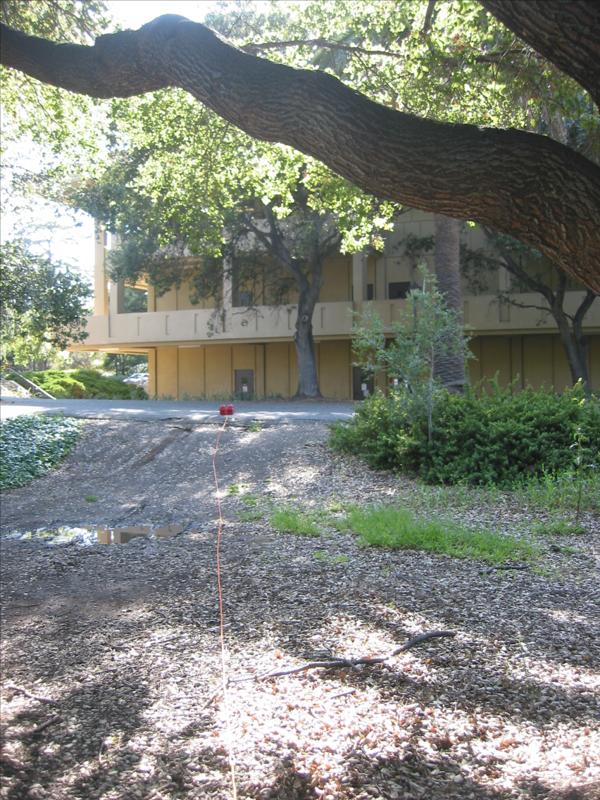}
\hspace*{\fill} 
\includegraphics[width=0.15\textwidth,height=100pt]{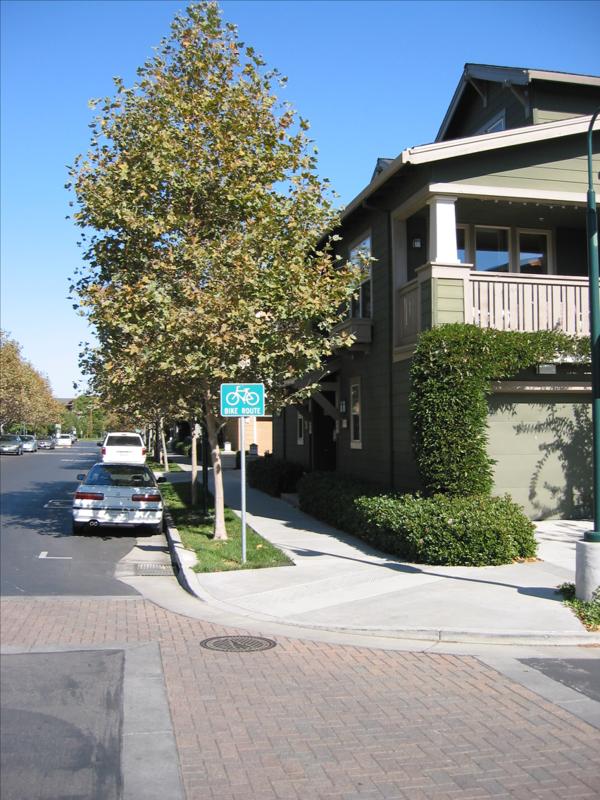}
\hspace*{\fill} 
\includegraphics[width=0.15\textwidth,height=100pt]{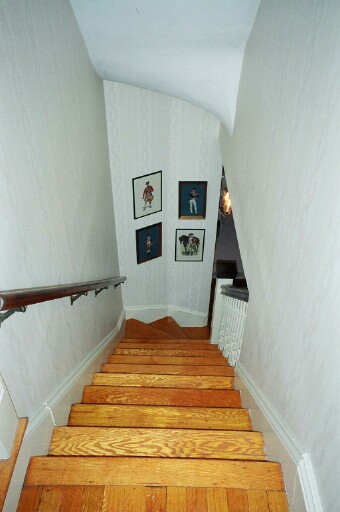}
\hspace*{\fill} 
\includegraphics[width=0.15\textwidth,height=100pt]{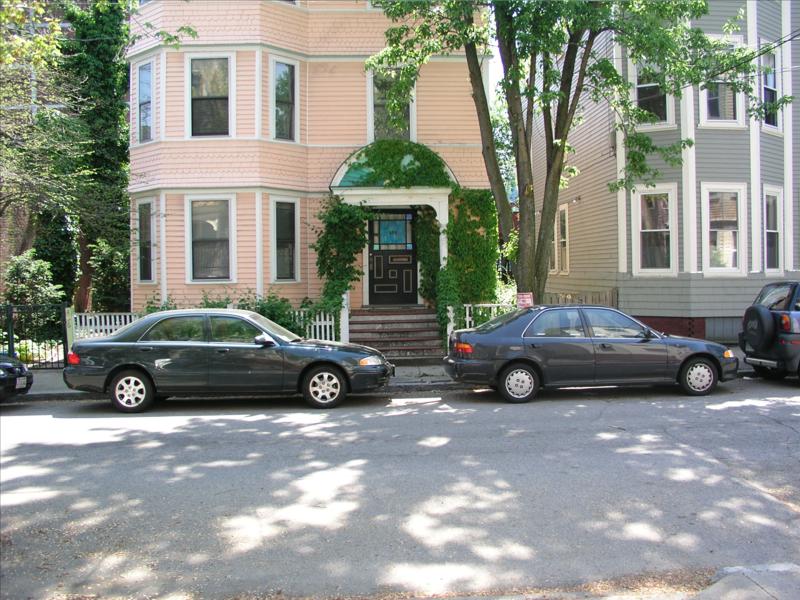}
\hspace*{\fill} 
\includegraphics[width=0.15\textwidth,height=100pt]{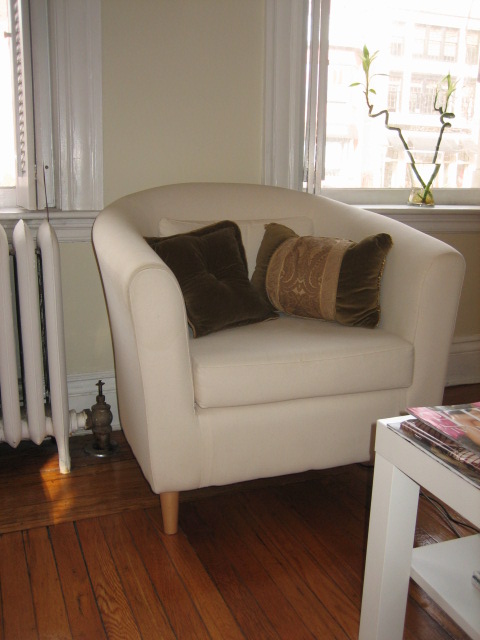}

\includegraphics[width=0.15\textwidth,height=100pt]{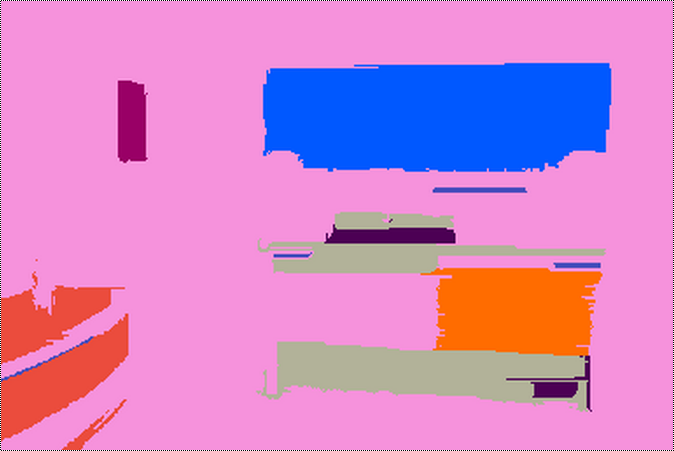}
\hspace*{\fill} 
\includegraphics[width=0.15\textwidth,height=100pt]{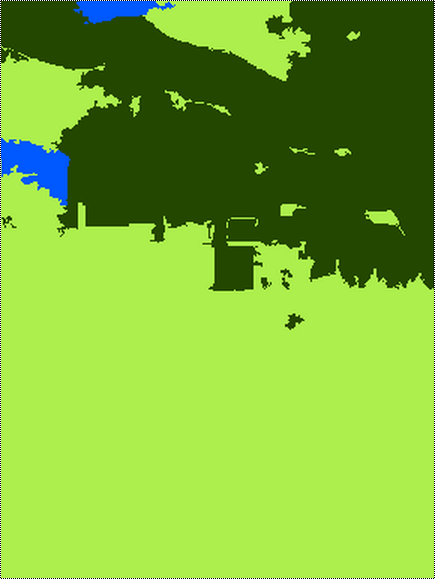}
\hspace*{\fill} 
\includegraphics[width=0.15\textwidth,height=100pt]{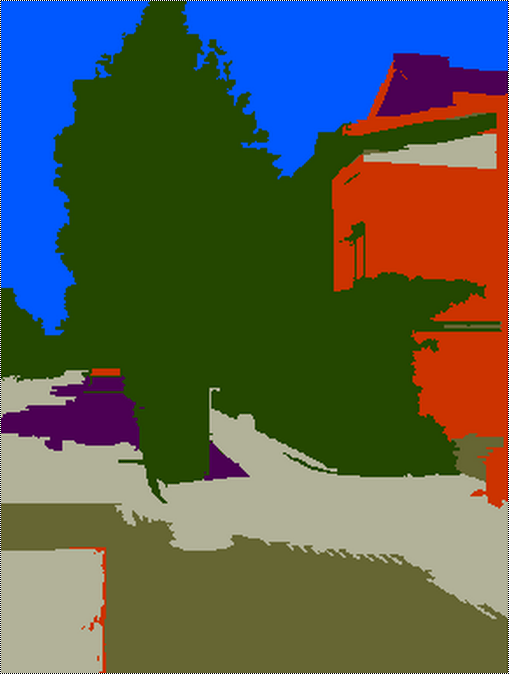}
\hspace*{\fill} 
\includegraphics[width=0.15\textwidth,height=100pt]{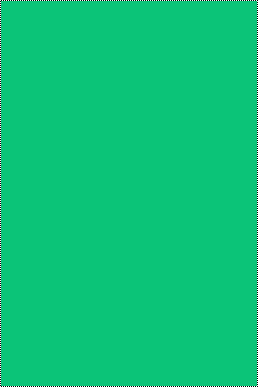}
\hspace*{\fill} 
\includegraphics[width=0.15\textwidth,height=100pt]{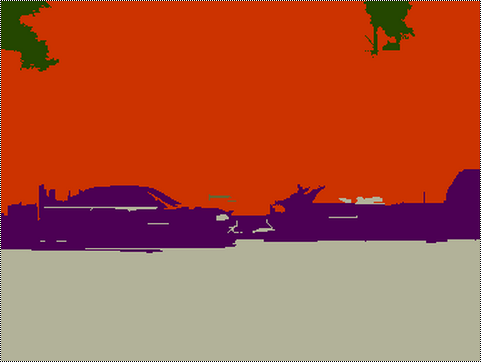}
\hspace*{\fill} 
\includegraphics[width=0.15\textwidth,height=100pt]{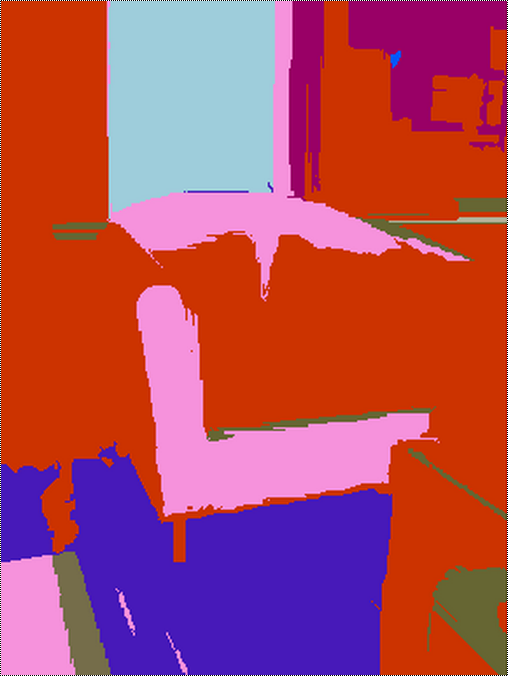}

\includegraphics[width=0.15\textwidth,height=100pt]{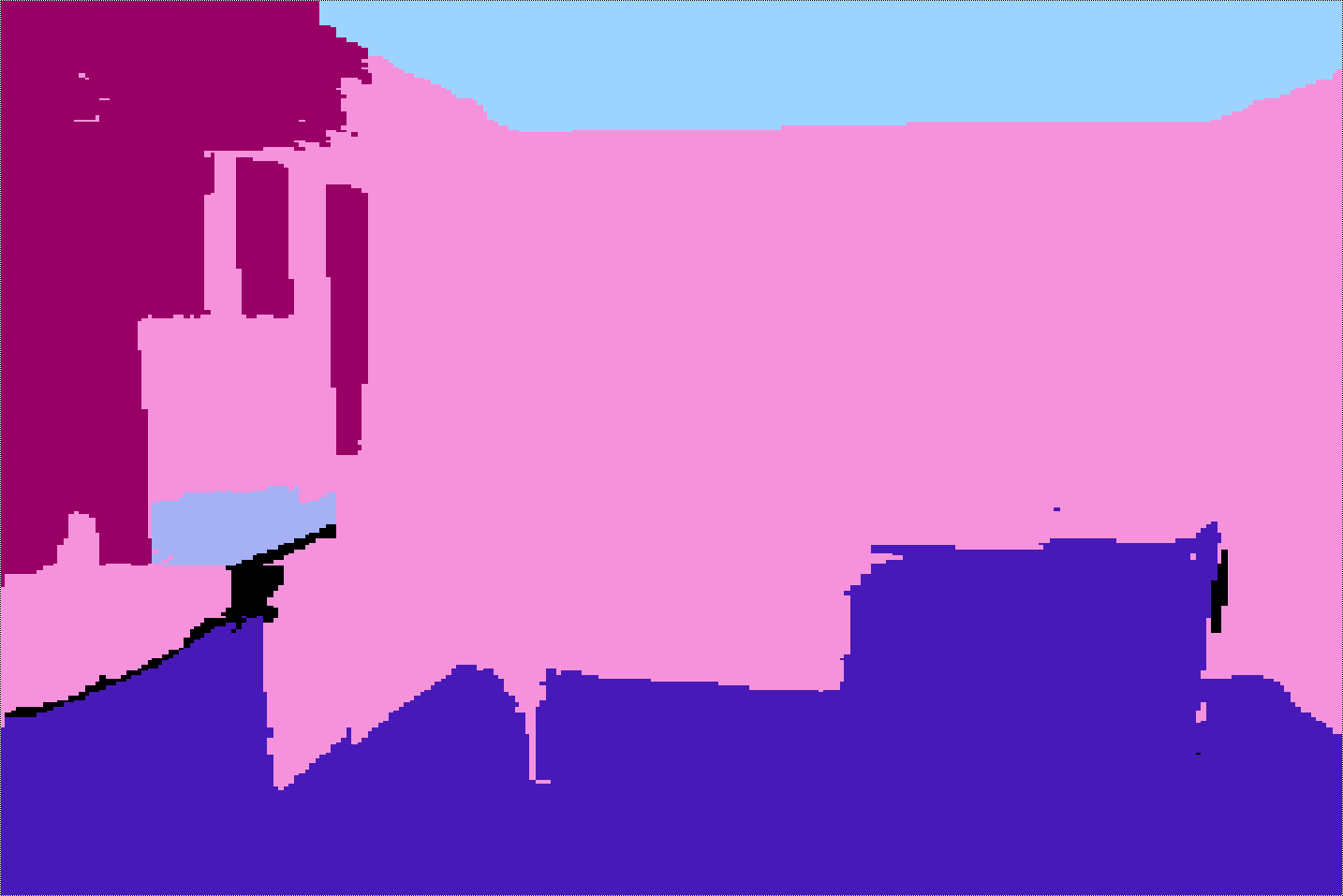}
\hspace*{\fill} 
\includegraphics[width=0.15\textwidth,height=100pt]{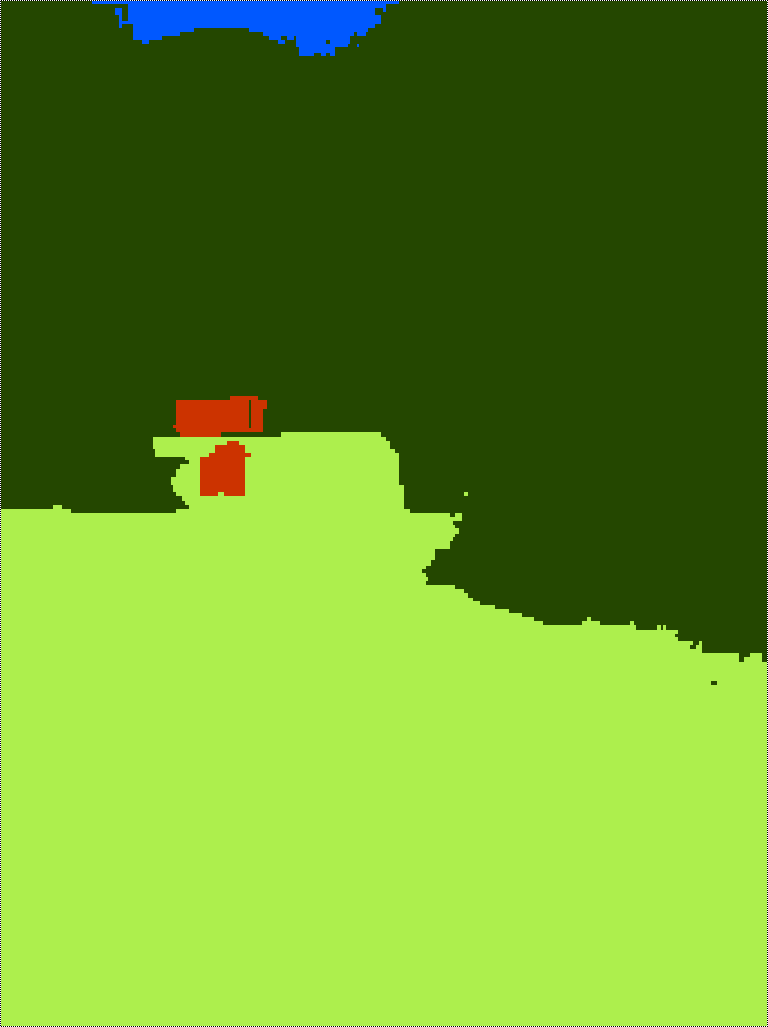}
\hspace*{\fill} 
\includegraphics[width=0.15\textwidth,height=100pt]{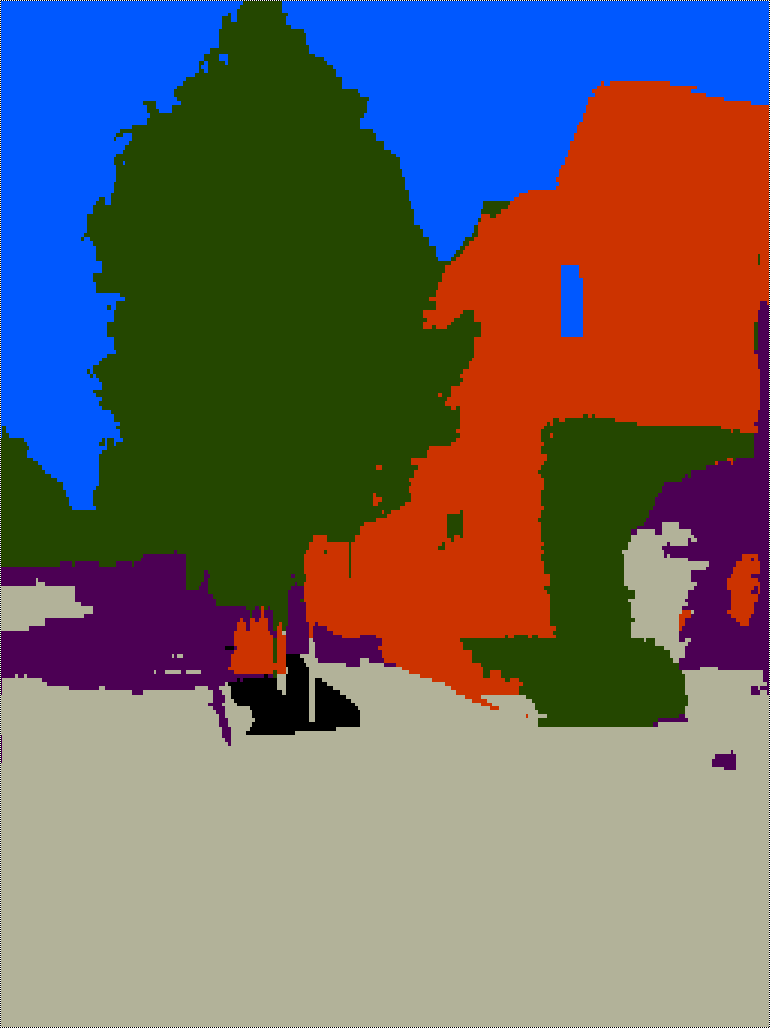}
\hspace*{\fill} 
\includegraphics[width=0.15\textwidth,height=100pt]{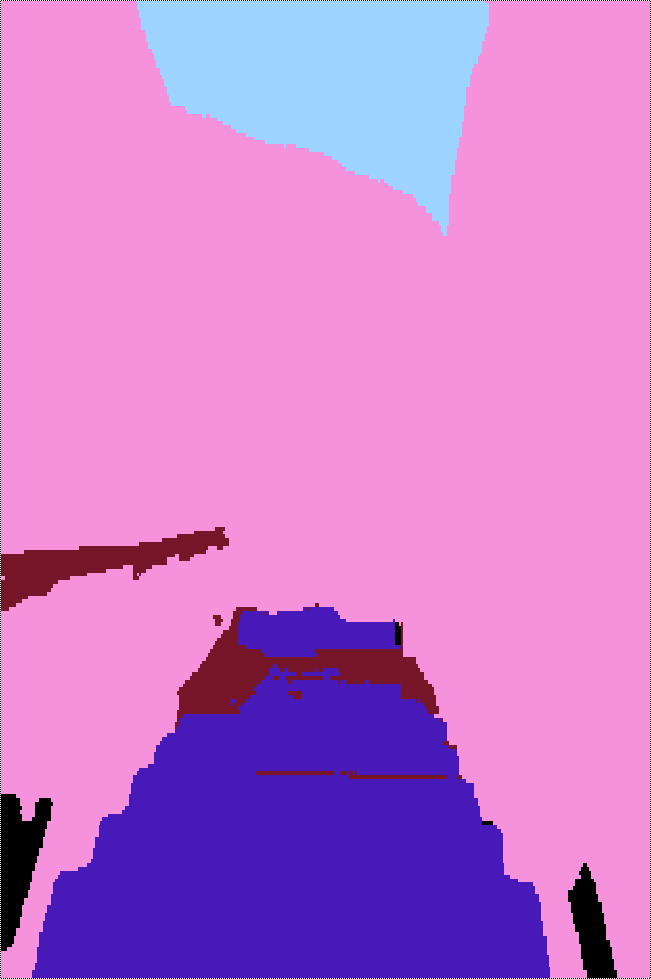}
\hspace*{\fill} 
\includegraphics[width=0.15\textwidth,height=100pt]{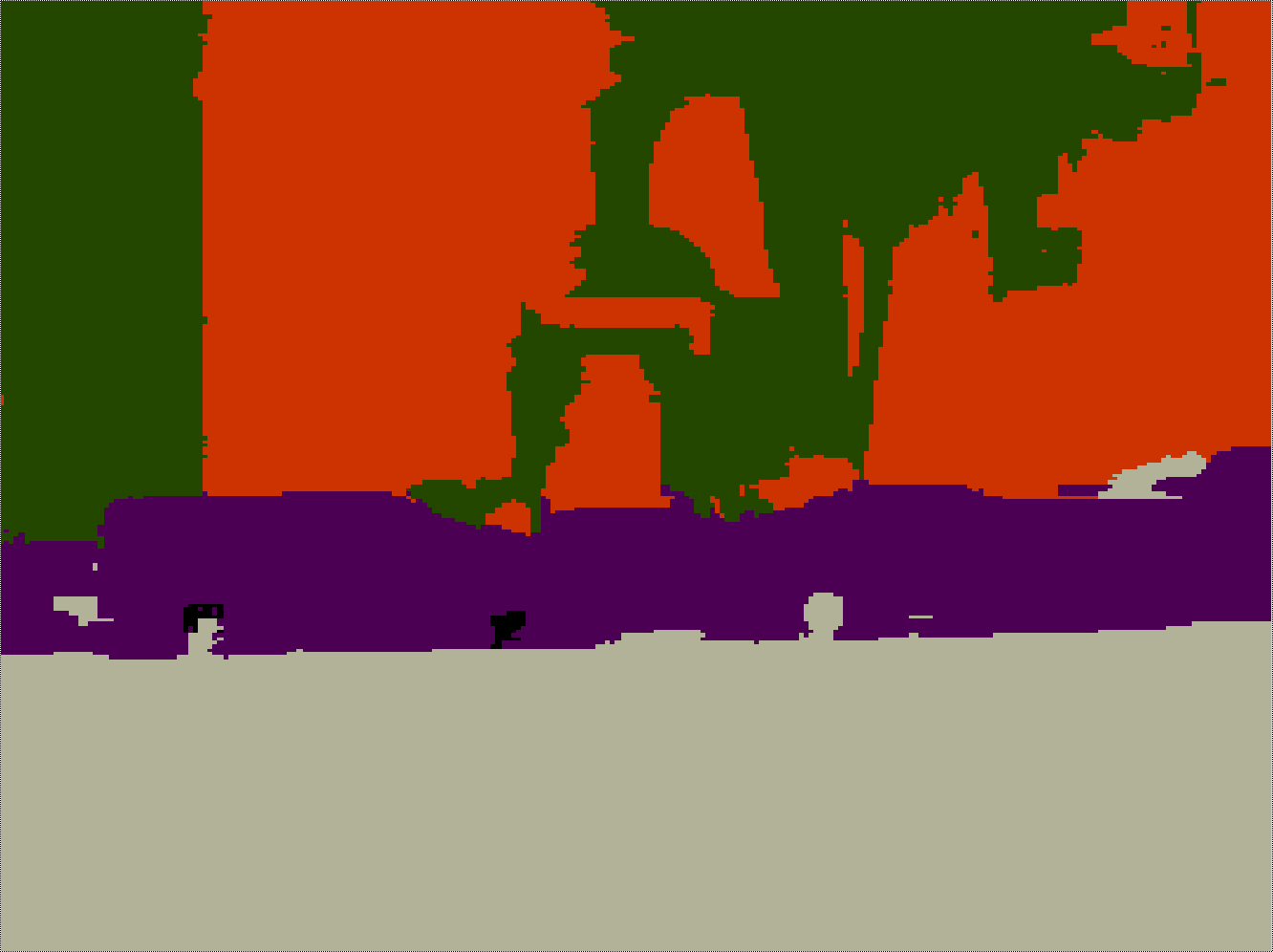}
\hspace*{\fill} 
\includegraphics[width=0.15\textwidth,height=100pt]{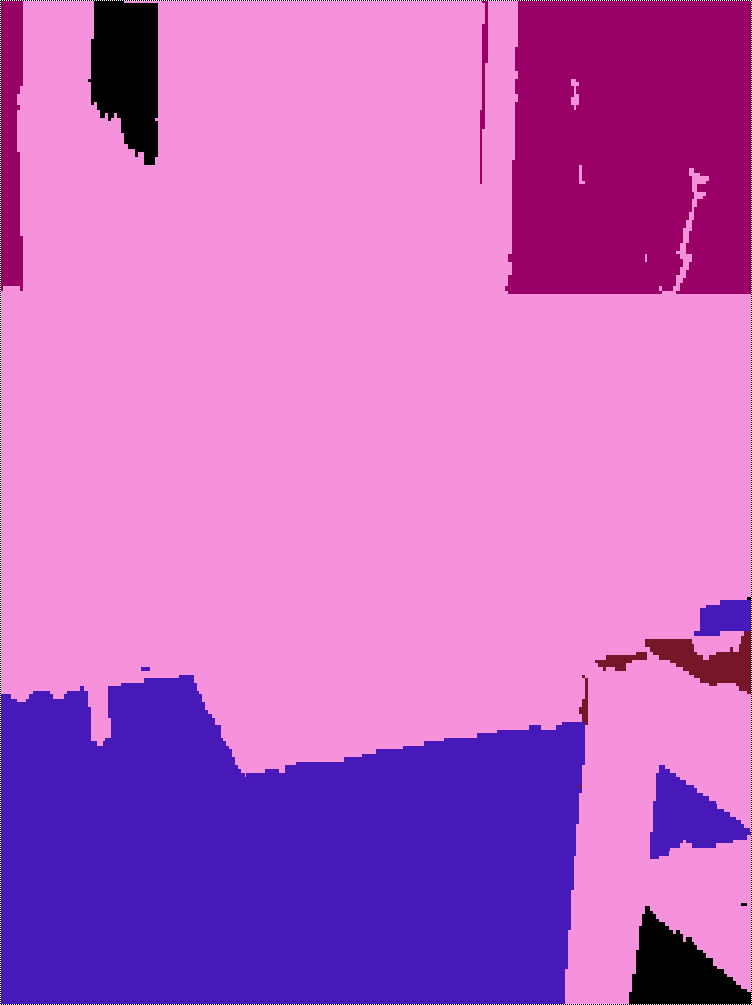}

\includegraphics[width=0.15\textwidth,height=100pt]{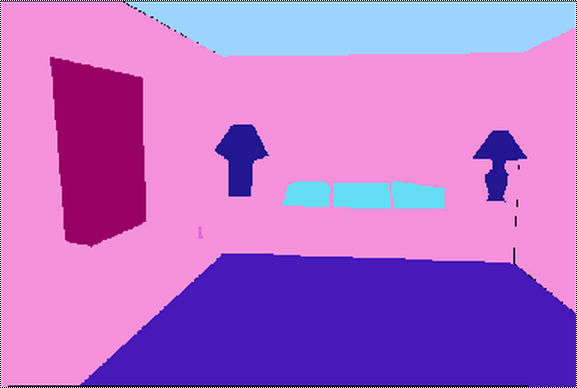}
\hspace*{\fill} 
\includegraphics[width=0.15\textwidth,height=100pt]{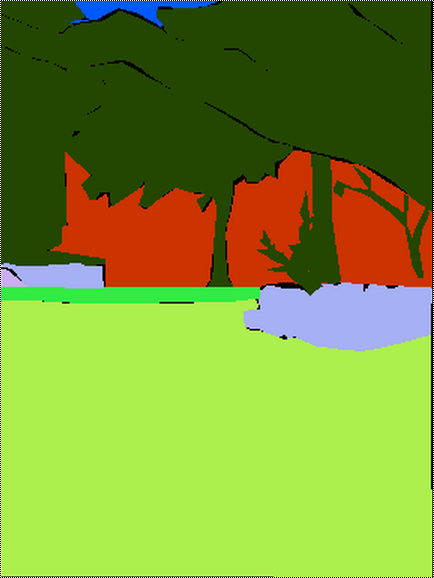}
\hspace*{\fill} 
\includegraphics[width=0.15\textwidth,height=100pt]{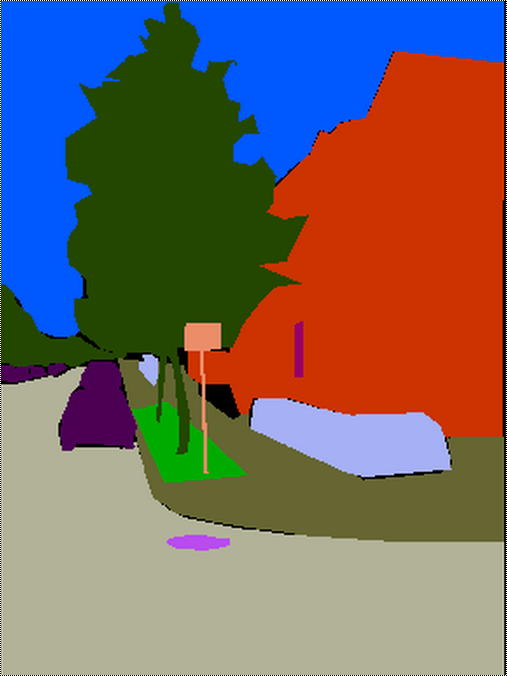}
\hspace*{\fill} 
\includegraphics[width=0.15\textwidth,height=100pt]{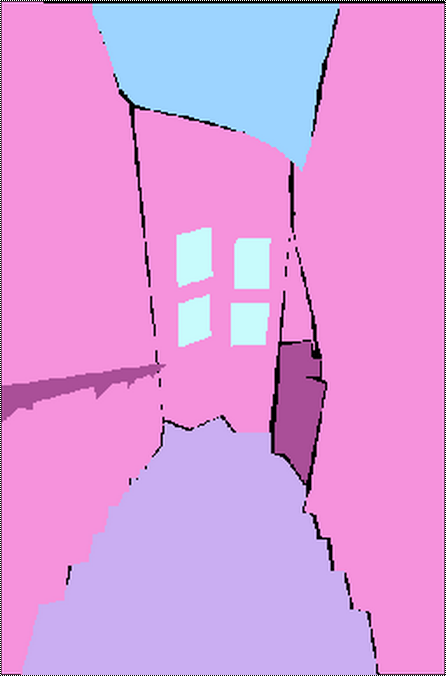}
\hspace*{\fill} 
\includegraphics[width=0.15\textwidth,height=100pt]{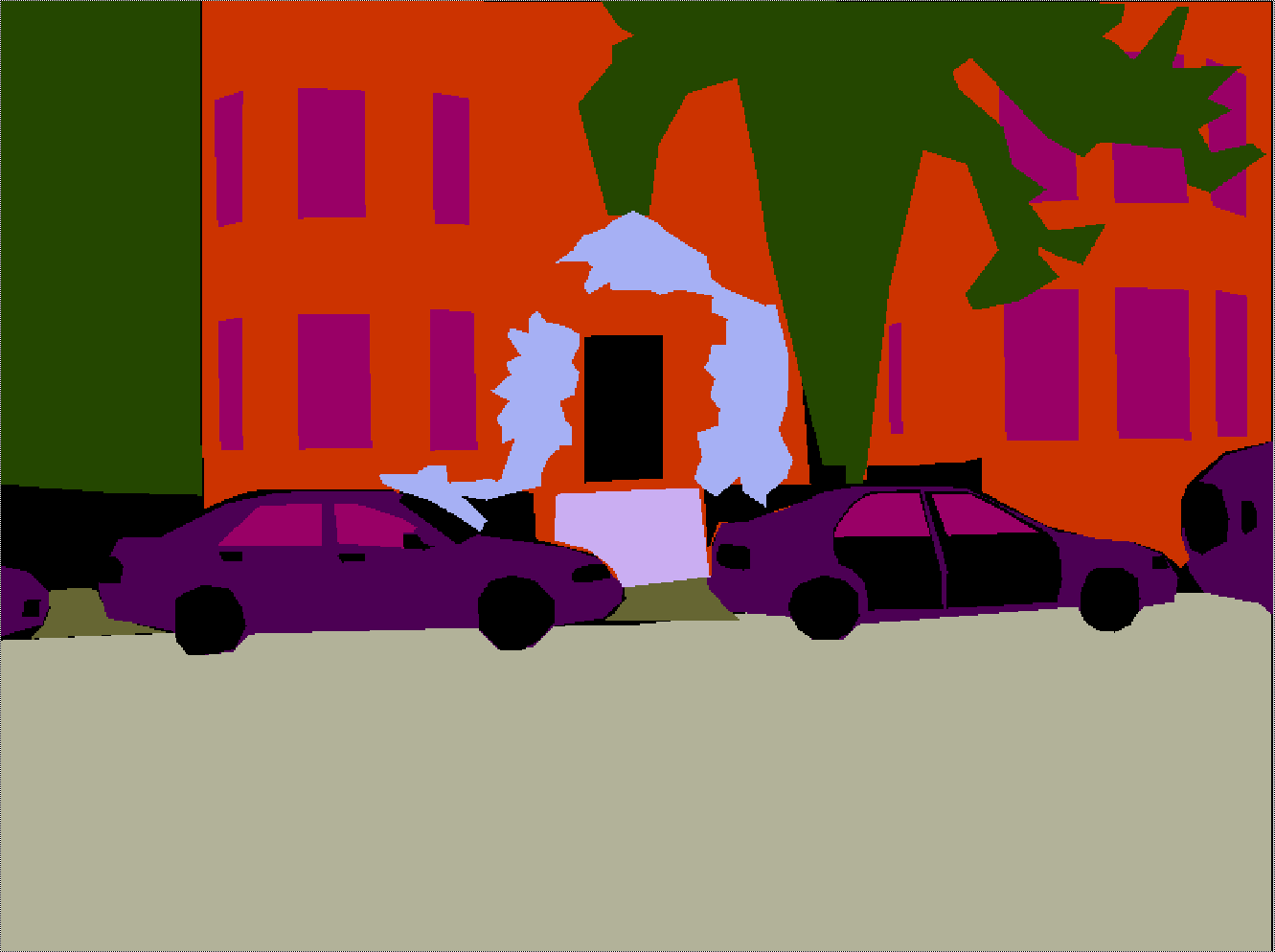}
\hspace*{\fill} 
\includegraphics[width=0.15\textwidth,height=100pt]{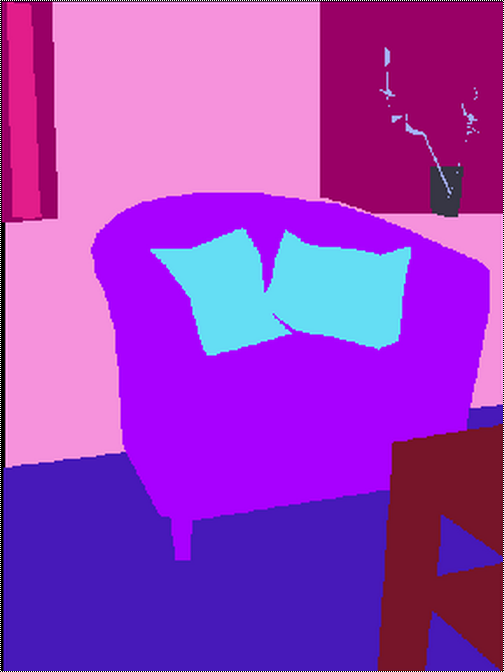}

\includegraphics[width=\textwidth]{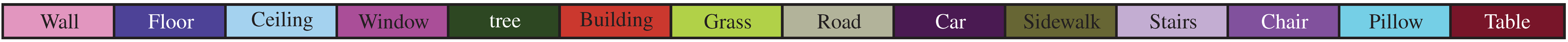}

\caption{Qualitative comparison of our results with those of Superparsing~\cite{Tighe:ijcv:2013} on LM-SUN. {\bf 1st row:} Query image; {\bf 2nd row:} Superparsing; {\bf 3rd row:} Our approach; {\bf 4th row:} Ground-truth.} \label{fig:lmsun}
\end{figure*}

In Table~\ref{table:lmsunmain}, we compare our results with those of~\cite{Tighe:ijcv:2013}, which constitutes the state-of-the-art on this dataset.
To the best of our knowledge, Superparsing~\cite{Tighe:ijcv:2013} is the only nonparametric approach that has been evaluated on this large-scale dataset. As a matter of fact, the scale of this dataset causes most nonparametric method to be intractable. By contrast, our efficient algorithm can still yield state-of-the-art accuracies in a reasonable time. 
In particular, our {\it sample \& filter} procedure takes 3.7 seconds per image on average, versus 13.1 seconds for Superparsing to transfer the labels. Furthermore, for each query image, our algorithm performs filtering on 367,080 superpixels on average, which is about 10 times larger than the 35,600 superpixels (200 retrieved images, each containing approximately 178 superpixels) analyzed by Superparsing. In other words, not only is our approach faster than Superparsing, but it can also exploit more labeled data.
Fig.~\ref{fig:lmsun} provides a qualitative comparison of our results with those of Superparsing.

\begin{table}[t!]\centering\renewcommand{\arraystretch}{1.2}
\caption{Comparison of our approach (Sample \& Filter) with Superparsing on LM-SUN. }
\vspace{-0.7cm}
\def\d{.1cm}
\def\r{30}
\small\addtolength{\tabcolsep}{2pt}
{
\begin{tabular}{lc|c}\\ 
 {}&{\bf \scriptsize per-pixel}&{\bf \scriptsize per-class}\\ \hline

{Sample \& Filter}&54.6&6.7\\

{Sample \& Filter (with CRF)}&55.1&6.6\\
\hline

{Superparsing~\cite{Tighe:ijcv:2013}}&{50.6}&{7.1}\\

{Superparsing (with MRF)~\cite{Tighe:ijcv:2013}}&{54.4}&{6.8}\\

\end{tabular}
}

\label{table:lmsunmain}
\end{table}


As in the previous section, we conducted an additional experiment using an ideal image ranking by making use of histograms of ground-truth annotations. Table \ref{table:lmsunideal} provides the results of this experiment. Note that, again, our approach has higher potential for improvement given a better image similarity measure.
\begin{table}[t!]\centering\renewcommand{\arraystretch}{1.2}
\caption{Comparison of our approach (Sample \& Filter) with Superparsing using an ideal image ranking on LM-SUN.}
\vspace{-0.7cm}
\def\d{.1cm}
\def\r{30}
\small\addtolength{\tabcolsep}{4pt}
{
\begin{tabular}{lc|c}\\ 
 {}&{\bf \scriptsize per-pixel}&{\bf \scriptsize per-class}\\ \hline

{Sample \& Filter (Ideal retrieval)}&69.3&15\\
\hline

{Superparsing~\cite{Tighe:ijcv:2013} (Ideal retrieval)}&{66}&{13.2}\\
\end{tabular}
}

\label{table:lmsunideal}
\end{table}

\section{Conclusion}
In this paper, we have introduced a nonparametric approach to scene parsing based on the concept of sampling and filtering. Instead of using a fixed retrieval set of images, our approach samples labeled superpixels, thus allowing us to obtain a more balanced set of data. This, in conjunction with our efficient filtering-based label transfer procedure, has proven effective at handling large-scale datasets. In particular, our approach has achieved accuracies that are competitive with the state-of-the-art nonparametric methods, while being faster than them. In the future, we intend to study better image similarity metrics, which, as evidenced by our analysis, has potential to further boost our accuracy.



{\small
\bibliographystyle{ieee}
\bibliography{egbib}

\begin{thebibliography}{10}\itemsep=-1pt

\bibitem{Adams:permutohedral:2010}
A.~Adams, J.~Baek, and M.~A. Davis.
\newblock Fast high-dimensional filtering using the permutohedral lattice.
\newblock {\em Computer Graphics Forum}, 2010.

\bibitem{Adams:2009:GKF}
A.~Adams, N.~Gelfand, J.~Dolson, and M.~Levoy.
\newblock Gaussian kd-trees for fast high-dimensional filtering.
\newblock {\em ACM Transactions on Graphics}, 28(3):21:1--21:12, 2009.

\bibitem{Barnes:2009:PAR}
C.~Barnes, E.~Shechtman, A.~Finkelstein, and D.~B. Goldman.
\newblock Patchmatch: A randomized correspondence algorithm for structural
  image editing.
\newblock {\em ACM Transactions on Graphics}, 28(3), 2009.

\bibitem{Barnes:2010:TGP}
C.~Barnes, E.~Shechtman, D.~B. Goldman, and A.~Finkelstein.
\newblock The generalized {PatchMatch} correspondence algorithm.
\newblock In {\em ECCV}, 2010.

\bibitem{Eigen:cvpr:2012}
D.~Eigen and R.~Fergus.
\newblock Nonparametric image parsing using adaptive neighbor sets.
\newblock In {\em CVPR}, 2012.

\bibitem{Farabet:CNL:2013}
C.~Farabet, C.~Couprie, L.~Najman, and Y.~LeCun.
\newblock Scene parsing with multiscale feature learning, purity trees, and
  optimal covers.
\newblock In {\em ICML}, 2012.

\bibitem{Gastal:2011:DTE}
E.~S.~L. Gastal and M.~M. Oliveira.
\newblock Domain transform for edge-aware image and video processing.
\newblock {\em ACM Transactions on Graphics}, 30(4):69:1--69:12, 2011.

\bibitem{Marian:Widerange:2015}
M.~George.
\newblock Image parsing with a wide range of classes and scene-level context.
\newblock In {\em CVPR}, 2015.

\bibitem{Gould:ICCV09}
S.~Gould, R.~Fulton, and D.~Koller.
\newblock Decomposing a scene into geometric and semantically consistent
  regions.
\newblock In {\em ICCV}, 2009.

\bibitem{Gould:eccv:2012}
S.~Gould and Y.~Zhang.
\newblock Patchmatchgraph: Building a graph of dense patch correspondences for
  label transfer.
\newblock In {\em ECCV}, 2012.

\bibitem{Kohli:2009:RHO}
P.~Kohli, L.~Ladick\'{y}, and P.~H. Torr.
\newblock Robust higher order potentials for enforcing label consistency.
\newblock {\em IJCV}, 82(3):302--324, 2009.

\bibitem{Kraehenbuehl:icml:2013}
P.~Kr{\"{a}}henb{\"{u}}hl and V.~Koltun.
\newblock Parameter learning and convergent inference for dense random fields.
\newblock In {\em ICML}, 2013.

\bibitem{Ladicky:pami:2013}
L.~Ladicky, C.~Russell, P.~Kohli, and P.~Torr.
\newblock Associative hierarchical random fields.
\newblock {\em PAMI}, 36(6):1056--1077, 2014.

\bibitem{Ladicky:IJCV:2013}
L.~Ladicky, C.~Russell, P.~Kohli, and P.~H. Torr.
\newblock Inference methods for crfs with co-occurrence statistics.
\newblock {\em IJCV}, 2013.

\bibitem{Liu:CVPR2015}
B.~Liu and X.~He.
\newblock Multiclass semantic video segmentation with object-level active
  inference.
\newblock In {\em CVPR}, 2015.

\bibitem{Liu:pami:2011}
C.~Liu, J.~Yuen, and A.~Torralba.
\newblock Nonparametric scene parsing via label transfer.
\newblock {\em PAMI}, 33(12):2368--2382, 2011.

\bibitem{Liu:siftflow:2011}
C.~Liu, J.~Yuen, and A.~Torralba.
\newblock Sift flow: Dense correspondence across scenes and its applications.
\newblock {\em PAMI}, 33(12):2368--2382, 2011.

\bibitem{long:fcn:2015}
J.~Long, E.~Shelhamer, and T.~Darrell.
\newblock Fully convolutional networks for semantic segmentation.
\newblock In {\em CVPR}, 2015.

\bibitem{Myeong:context:2012}
H.~Myeong, J.~Y. Chang, and K.~M. Lee.
\newblock Learning object relationships via graph-based context model.
\newblock In {\em CVPR}, 2012.

\bibitem{Myeong:tensor:2013}
H.~Myeong and K.~M. Lee.
\newblock Tensor-based high-order semantic relation transfer for semantic scene
  segmentation.
\newblock In {\em CVPR}, 2013.

\bibitem{Oliva:2001:GIST}
A.~Oliva and A.~Torralba.
\newblock Modeling the shape of the scene: A holistic representation of the
  spatial envelope.
\newblock {\em IJCV}, 42(3):145--175, 2001.

\bibitem{Sharma:DeepParsing:2015}
A.~Sharma, O.~Tuzel, and D.~W. Jacobs.
\newblock Deep hierarchical parsing for semantic segmentation.
\newblock In {\em CVPR}, 2015.

\bibitem{Shotton:textonboost:eccv}
J.~Shotton, J.~Winn, C.~Rother, and A.~Criminisi.
\newblock Textonboost: Joint appearance, shape and context modeling for
  mulit-class object recognition and segmentation.
\newblock In {\em ECCV}, 2006.

\bibitem{Shuai:paramAndNonparam:2015}
B.~Shuai, G.~Wang, Z.~Zuo, B.~Wang, and L.~Zhao.
\newblock Integrating parametric and non-parametric models for scene labeling.
\newblock In {\em CVPR}, 2015.

\bibitem{Singh:cvpr:2013}
G.~Singh and J.~Kosecka.
\newblock Nonparametric scene parsing with adaptive feature relevance and
  semantic context.
\newblock In {\em CVPR}, 2013.

\bibitem{Tighe:Finding:2015}
J.~Tighe and S.~Lazebnik.
\newblock Finding things: Image parsing with regions and per-exemplar
  detectors.
\newblock In {\em CVPR}, 2013.

\bibitem{Tighe:ijcv:2013}
J.~Tighe and S.~Lazebnik.
\newblock Superparsing: Scalable nonparametric image parsing with superpixels.
\newblock {\em IJCV}, 101(2):329--349, 2013.

\bibitem{Tung:Collage:eccv}
F.~Tung and J.~J. Little.
\newblock Collageparsing: Nonparametric scene parsing by adaptive overlapping
  windows.
\newblock In {\em ECCV}, 2014.

\bibitem{Wong:1980:datasample}
C.~K. Wong and M.~C. Easton.
\newblock An efficient method for weighted sampling without replacement. siam
  journal of computing.
\newblock {\em SIAM Journal o Computing}, 9(1):111--113, 1980.

\bibitem{sunDatabase:cvpr:2010}
J.~Xiao, J.~Hays, K.~Ehinger, A.~Oliva, and A.~Torralba.
\newblock Sun database: Large-scale scene recognition from abbey to zoo.
\newblock In {\em CVPR}, 2010.

\bibitem{Yang:rare:2015}
J.~Yang, B.~Price, S.~Cohen, and M.-H. Yang.
\newblock Context driven scene parsing with attention to rare classes.
\newblock In {\em CVPR}, 2014.

\end{thebibliography}
}

\end{document}